\pdfoutput=1

\documentclass[11pt]{article}
\usepackage[]{acl}

\usepackage{times}
\usepackage{latexsym}
\usepackage{lipsum}
\usepackage{booktabs}
\usepackage[labelfont=bf]{caption}
\usepackage{multirow}
\usepackage{graphicx}
\usepackage{amssymb}
\usepackage{amsmath}
\usepackage{soul}
\usepackage[T1]{fontenc}

\usepackage[utf8]{inputenc}

\usepackage{microtype}

\usepackage{inconsolata}

\usepackage{graphicx} 
\usepackage{natbib}  
\usepackage{caption} 
\usepackage{multirow}
\usepackage{algorithm}
\usepackage{algpseudocode}
\usepackage{amsmath}
\usepackage{booktabs}
\usepackage{amsfonts}
\usepackage{blkarray, bigstrut}
\parskip=\medskipamount
\usepackage{amssymb}
\usepackage{pifont}
\newcommand{\vicunavonefive}{{\tt vicuna-7b-v1.5}}

\newcommand{\zephyrbeta}{{\tt zephyr-7b-beta}}

\newcommand{\fl}{{\tt fluency}}
\newcommand{\coh}{{\tt coherence}}
\newcommand{\rel}{{\tt relevance}}
\newcommand{\fa}{{\tt faithfulness}}
\newcommand{\asp}{{\tt aspect coverage}}
\newcommand{\sent}{{\tt sentiment consistency}}
\newcommand{\spec}{{\tt specificity}}
\newcommand{\cl}{{\tt clarity}}
\newcommand{\ifo}{{\tt informativeness}}
\newcommand{\foa}{{\tt format adherence}}
\newcommand{\qr}{{\tt query relevance}}

\newcommand{\gemmaone}{{\tt Gemma-1.1-7b-it}}
\newcommand{\llamainstruct}[2]{{\tt LLaMA-$#1$.$#2$B-Instruct}}
\newcommand{\mistralinstructv}[2]{{\tt Mistral-$#1$B-Instruct-v$#2$}}
\newcommand{\qweninstruct}[1]{{\tt Qwen$#1$-7B-instruct}}
\newcommand{\gemmatwo}{{\tt Gemma-2-9b-it}}
\newcommand{\mixtral}{{\tt Mixtral-8x7B-Instruct-v0.1}}
\newcommand{\gptfour}{{\tt GPT-4o}}
\newcommand{\gptfourstandard}{{\tt GPT-4}}

\newcommand{\qweninstructnew}{{\tt Qwen/Qwen2-7B-Instruct}}

\newcommand{\mixtralnew}{{\tt mistralai/Mixtral-8x22B-Instruct-v0.1}}

\newcommand{\gptturbo}{{\tt GPT-3.5-Turbo}}

\usepackage[labelfont=bf]{caption}
\usepackage{multirow}
\newcommand*\rot{\rotatebox{90}}
\usepackage{newfloat}
\usepackage{listings}

\usepackage{xcolor}
\usepackage[capitalise,noabbrev]{cleveref}
\usepackage{hyperref}

\newcommand\blfootnote[1]{%
  \begingroup
  \renewcommand\thefootnote{}\footnote{#1}%
  \addtocounter{footnote}{-1}%
  \endgroup
}

\title{"This Suits You the Best": Query Focused Comparative Explainable Summarization}

\author{Arnav Attri$^\diamondsuit$$^\clubsuit$,
Anuj Attri$^\diamondsuit$$^\clubsuit$, Pushpak Bhattacharyya$^\clubsuit$ 
\\ 
\textbf{Suman Banerjee$^\mathcal{F}$, Amey Patil$^\mathcal{F}$, Muthusamy Chelliah$^\mathcal{F}$, Nikesh Garera$^\mathcal{F}$}\\
\textbf{}
        $^\clubsuit$Computer Science and Engineering, IIT Bombay, India, 
        $^\mathcal{F}$Flipkart, India \\
        \texttt{\{arnavcs, ianuj,
        pb\}@cse.iitb.ac.in}
        }

\begin{document}
\maketitle
\blfootnote{$^\diamondsuit$ Equal contribution}
\begin{abstract}
Product recommendations inherently involve comparisons, yet traditional opinion summarization often fails to provide holistic comparative insights. We propose the novel task of generating \textbf{Q}uery-\textbf{F}ocused \textbf{C}omparative \textbf{E}xplainable \textbf{S}ummaries (\textsc{QF-CES}) using \textbf{M}ulti-Source \textbf{O}pinion \textbf{S}ummarization (\textsc{M-OS}). To address the lack of query-focused recommendation datasets, we introduce \textsc{MS-Q2P}, comprising $\mathbf{7,500}$ queries mapped to $\mathbf{22,500}$ recommended products with metadata. We leverage Large Language Models (LLMs) to generate tabular comparative summaries with query-specific explanations. Our approach is personalized, privacy-preserving, recommendation engine-agnostic, and category-agnostic. \textsc{M-OS} as an intermediate step reduces inference latency approximately by $\mathbf{40\%}$\footnote{This percentage reflects the average time reduction across $50$ distinct summaries, each generated $50$ times for reliability. \textsc{M-OS} averaged $9.99$ seconds per summary, compared to $16.55 $seconds for DIA.} compared to the direct input approach (\textsc{DIA}), which processes raw data directly. We evaluate open-source and proprietary LLMs for generating and assessing \textsc{QF-CES}. Extensive evaluations using \textsc{\textbf{QF-CES-PROMPT}} across $5$ dimensions (\texttt{clarity}, \texttt{faithfulness}, \texttt{informativeness}, \texttt{format adherence}, and \texttt{query relevance}) showed an average Spearman correlation of $\mathbf{0.74}$ with human judgments, indicating its potential for \textsc{QF-CES} evaluation.
\end{abstract}

\section{Introduction}
\begin{figure}[t]
    \centering
    \includegraphics[width=1\columnwidth]{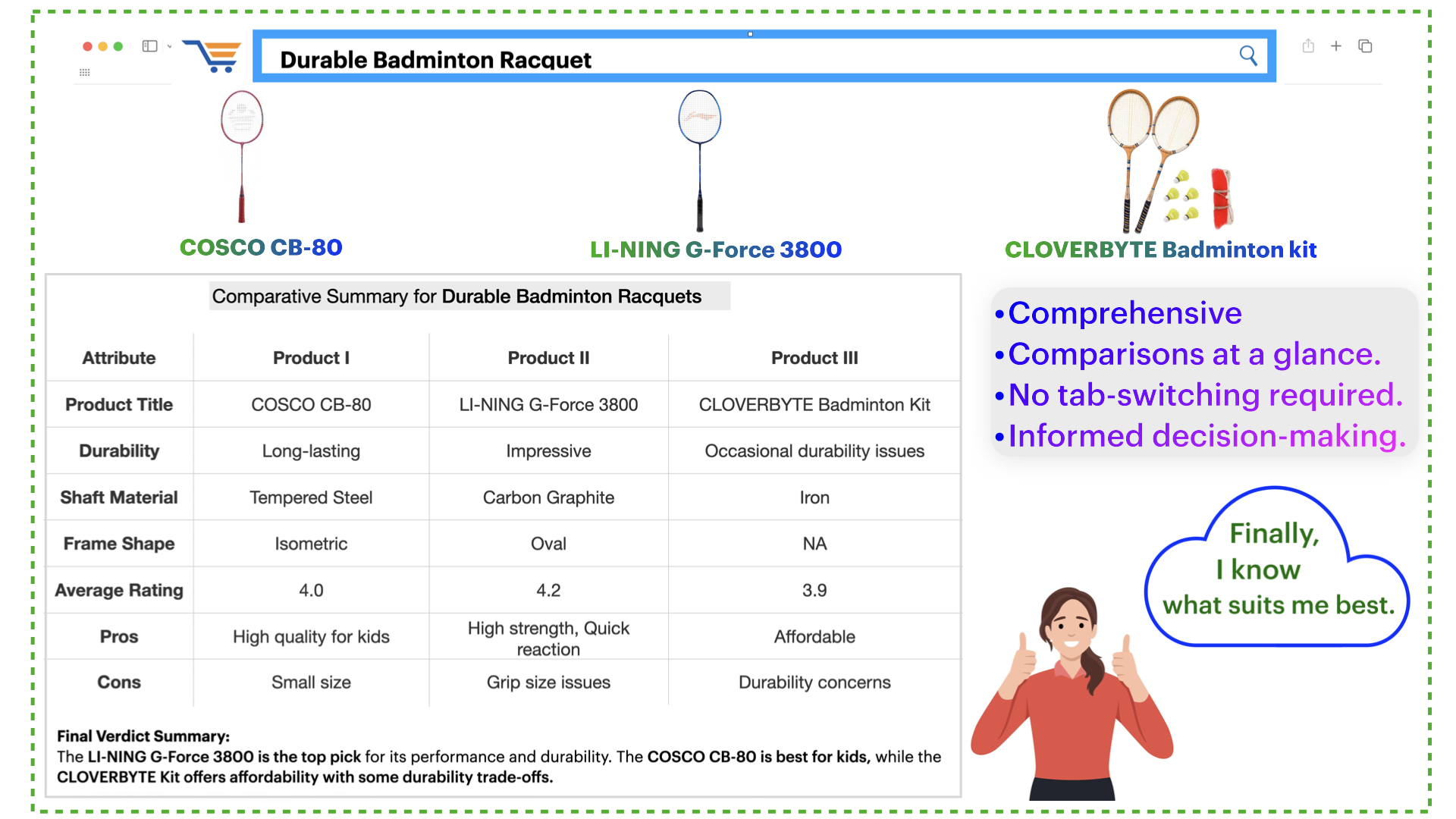}
    \caption{\textsc{QF-CES} enables quick comparison of top-3 recommended products for confident decisions without tab-switching. Check Figure \ref{fig:big} for details.} 
    \label{fig:first}
\end{figure}

E-commerce platforms host a vast array of products, but users face challenges in decision-making despite recommendation systems. Users, each with unique quality preferences, budget constraints, and desired features, often find themselves sifting through specifications and reviews of multiple (but often quite similar) products.  While recommendation systems, match products to queries, they lack comparative insights crucial for informed decisions. Users struggle to understand how recommended items stack up against each other in ways that matter most to their individual needs and specific queries.
\citet{tay-2019} highlight that opinion summarization approaches condense reviews by frequently emphasizing recurring aspects, potentially introducing bias and giving users a skewed perception of their importance. However, this approach overlooks valuable information embedded within product metadata, highlighting the need for \textsc{M-OS}. \citet{im, LI} explored methods incorporating reviews, images, and metadata to provide users with informative summaries that capture both subjective opinions and objective product attributes, as demonstrated by \citet{TJ-MOS}.
These approaches generate single-product summaries without user query context or cross-product comparisons, forcing users to manually compare items, leading to decision fatigue and a sub-optimal shopping experience.

\begin{figure*}[htp]
    \centering
    \includegraphics[width=2\columnwidth]{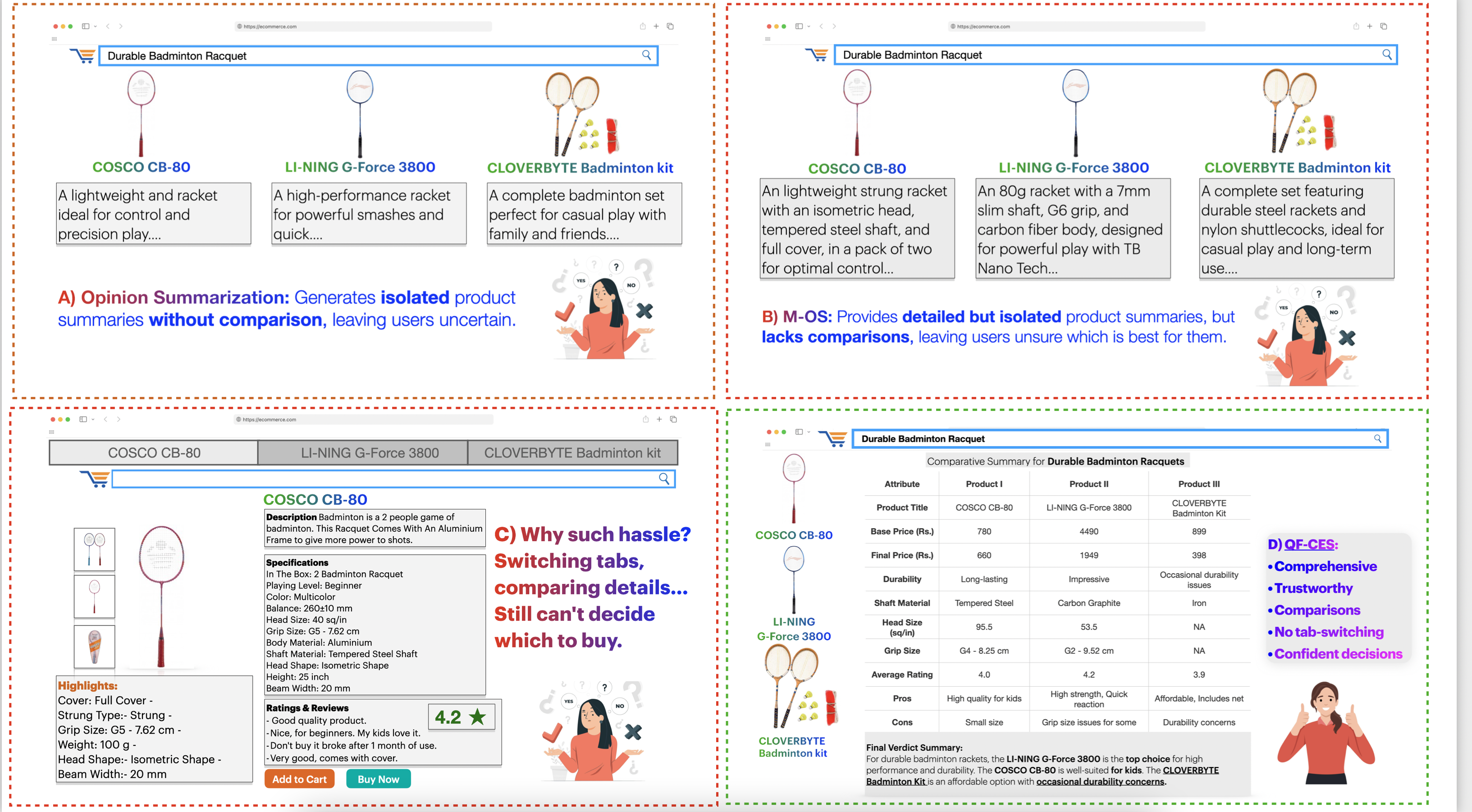}
    \caption{Comparison of approaches: (A) Traditional opinion summaries, (B) M-OS, (C) Single-product views with tab navigation, and (D) textsc{QF-CES}. Unlike traditional methods with isolated summaries, textsc{QF-CES} offers side-by-side comparisons and a final verdict, eliminating tab-switching and enhancing decision-making confidence.}
    \label{fig:big}
\end{figure*}

We propose Query-Focused Comparative Explainable Summarization \textsc{QF-CES} to address these limitations. \textsc{QF-CES} provides targeted, comparative insights for recommended products in one place, as shown in Figure \ref{fig:big}, facilitating informed decision-making. It generates a comparative summary in a tabular format, complemented by a Natural Language Explanation (NLE) as a final verdict that directly addresses the user's query.

\underline{\textbf{Problem Statement:}} 

\textbf{Input:} Query and top-$k$ $(k=3)$ recommended products

\textbf{Output:} \textsc{QF-CES} with tabular comparison and final verdict explanation.

Our contributions are:
\begin{enumerate}
\item \textbf{\textsc{QF-CES:}} A novel task using LLMs to generate Query-Focused Comparative Explainable Summaries. It leverages Multi-Source Opinion Summaries (M-OS) as an intermediate step, reducing inference latency by $\mathbf{40\%}$ (Section \ref{time}) compared to raw data input.
\item \textbf{\textsc{MS-Q2P:}} A new dataset featuring queries with top-3 recommended products and associated metadata (Section \ref{q2p_dataset}).
\item \textbf{\textsc{CES-EVAL:}}  An \textsc{QF-CES} evaluation benchmark dataset (Section \ref{annotation_dataset}), with $2,500$ summary annotations, assessing $10$ comparative summaries for $50$ queries from the \textsc{MS-Q2P}. The evaluation covers $\mathbf{5}$ \textbf{dimensions}- \texttt{clarity}, \texttt{faithfulness}, \texttt{informativeness}, \texttt{format adherence}, and \texttt{query relevance} (Appendix \ref{cex_metrics}).
\item \textbf{\textsc{QF-CES-PROMPT:}} A set of dimension-dependent prompts enables comparative summary generation and evaluation of all the aforementioned $5$ dimensions.  To the best of our knowledge, we are the first to create a structured tabular comparison with a final verdict summary that directly addresses the user’s specific query.
\item Benchmarking of $9$ recent LLMs (closed and open-source) on the aforementioned $5$ dimensions for the task of comparative summaries, which to the best of our knowledge is first of its kind (Table \ref{tab:human_cex_score_round2}, Section \ref{results_analysis}).
\item Detailed analysis, comparing an open-source LLM against $\mathbf{4}$ closed-source LLMs as evaluators for automatic evaluation of comparative summaries on $5$ dimensions. Analysis indicates that \textsc{QF-CES-PROMPT} emerges as a good alternative for reference-free evaluation of comparative summaries showing a high Spearman correlation of $\mathbf{0.74}$ on average with humans (Table \ref{tab:main_results_table}). 
\end{enumerate}

\section{Related Work}
\textbf{Explainable Recommendation} has been an active area of research in recent years, with early contributions from \citet{chen2018neural} and \citet{wang2018explainable}. \citet{li2020generate} and \citet{yang2021explanation} furthered the field, leading to \textsc{PETER}, a personalized transformer for explainable recommendation by \citet{peter2023peter}. \citet{colas2023knowrec} introduced \textsc{KnowRec}, a knowledge-grounded model, and \citet{gcre2023} enhanced explanations by extracting comparative relation tuples. \citet{dre2023} aligned LLMs for recommendation explanations, and \citet{uncertainty2023} leveraged LLMs to generate explanations. (\citet{Ji}, \citet{count}, \citet{LiLi}) Generate templatized explanations using item attributes and sentiment from reviews. 

\textbf{Comparative Summarization} has received limited attention. \citet{collaborative2022} generated contrastive summaries and a common summary from user reviews, \citet{comparative2022} review-based explanations for recommended items, \citet{echterhoff2023} generated aspect-aware comparative sentences, while \citet{constraints2021} proposed a framework incorporating comparative constraints into recommendation models. 

\textbf{LLM-based Evaluators} Traditional metrics like ROUGE \citep{ro} and BLEU \citep{bleu} often misalign with human judgments for opinion summaries. Recent NLP advancements, particularly in LLMs, offer promising alternatives. Studies have explored LLM-based evaluation methods \citep{fu, chiang-lee-2023-large, closer, wanggpt, kocmi-federmann-2023-large}, including Chain of Thought approaches \citep{liuug, wei2023} and reference-free evaluation \citep{chianggg}. \citet{op} proposed two prompt strategies for opinion summary evaluation on $7$ metrics. 

Our work differs from the existing work through (\textit{1}) \textbf{Consolidated Comparison} of three products simultaneously; (\textit{2}) \textbf{Query-Based Personalization}, preserving privacy; (\textit{3}) \textbf{Dynamic Attribute Generation} tailored to user queries; (\textit{4}) \textbf{Category-Agnostic} approach applicable across product domains; (\textit{5}) \textbf{Recommendation-Engine Agnostic}, functioning with any ranking system; and (\textit{6}) \textbf{Multi-Source Integration}, generating comprehensive summaries beyond user reviews. These features collectively offer a more versatile, privacy-conscious, and informative comparative summarization solution.

\section{Methodology}\label{method} Our study investigates LLMs' capabilities in generating and evaluating comparative summaries, an underexplored area, despite their success in various NLG tasks. We leverage the \textsc{MS-Q2P} dataset (Section \ref{q2p_dataset}),  enabling a thorough assessment of LLMs in comparative summarization.

\subsection{Multi-Source Opinion Summary (M-OS)}
We developed \textsc{M-OS} using an LLM ensemble, integrating diverse product attributes including product title, descriptions, key features, specifications, reviews, and average ratings for comprehensive representation. This approach establishes a robust foundation for \textsc{QF-CES} generation while reducing inference latency (Section \ref{time}). Our prompting-based methodology avoids fine-tuning overhead, offering an efficient solution. We evaluate \textsc{M-OS} quality by adapting the \textsc{OP-PROMPT} framework \citet{op} across 7 dimensions (\texttt{fluency}, \texttt{coherence}, \texttt{relevance}, \texttt{faithfulness}, \texttt{aspect coverage}, \texttt{sentiment consistency}, and \texttt{specificity}),  (Section \ref{mos_eval}), identifying the top-performing LLM (Table \ref{tab:opsum}) for subsequent \textsc{QF-CES} production.
\begin{figure*}[htp]
    \centering
    \includegraphics[width=2\columnwidth]{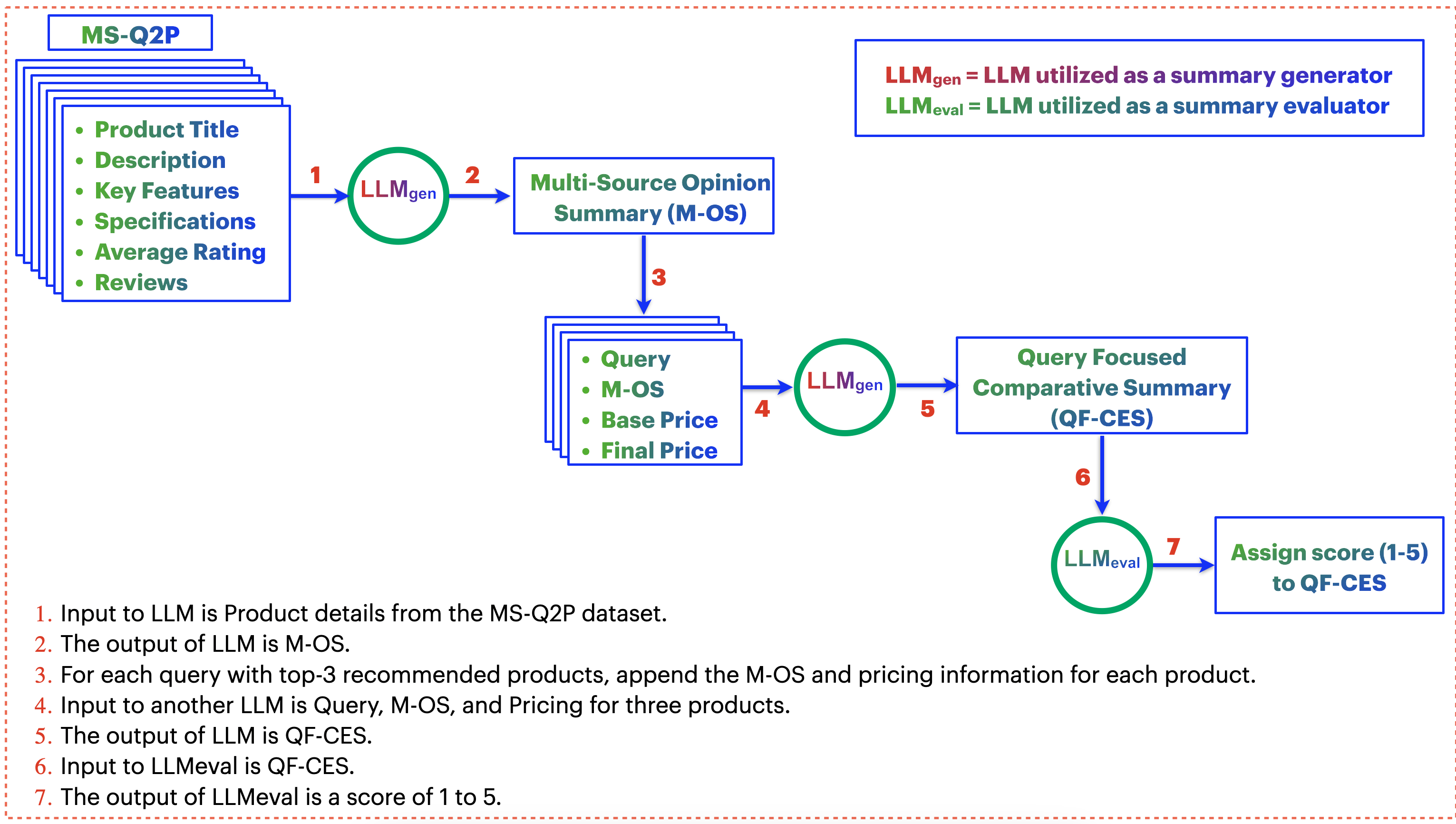}
    \caption{A Multi-phase Pipeline for Generating \textsc{QF-CES} using \textsc{M-OS} and Large Language Models (LLMs). The pipeline involves using LLMs both as summary generators (LLMgen) and summary evaluators (LLMeval) to create and assess \textsc{QF-CES} across various dimensions, incorporating product details from the \textsc{MS-Q2P} dataset. The points $1$ through $7$ describe the flow of inputs and outputs between the LLMs, from generating \textsc{M-OS} to evaluating \textsc{QF-CES}.}
    \label{fig:arch}
\end{figure*}
\subsection{Query-Focused Comparative Explaniable Summary (QF-CES)}
\textsc{QF-CES} generation utilizes \textsc{M-OS} that achieved the highest average score across all $7$ dimensions. Our approach uses sophisticated, prompt engineering, guiding the LLM through a detailed, step-by-step process. The LLM assigned an expert role, dynamically selects relevant product attributes based on user queries, product specifics, and attribute importance. The resulting \textsc{QF-CES} presents a tabular comparison of top-$k$ ($k=3$) recommended products, including titles, prices, ratings, selected attributes, and user-derived pros/cons. Missing attributes are marked \textit{"NA"}. A concise explanation provides a final verdict that directly addresses the user's query-specific needs, providing a personalized, query-focused comparison tailored to the user's requirements. 

\subsection{Evaluation of QF-CES} 
To the best of our knowledge, we are the first to present an evaluation of comparative summaries using reference-free metrics with both open- and closed-source LLMs as evaluators. Our \textsc{QF-CES-PROMPT} introduces metric-dependent prompts assessing generated \textsc{QF-CES} across $5$ dimensions: \texttt{clarity}, \texttt{faithfulness}, \texttt{informativeness}, \texttt{format adherence}, and \texttt{query relevance}. These dimension-specific prompts are applied to various LLMs for \textsc{QF-CES} evaluation. Additionally, we introduce \textsc{CES-EVAL}, a benchmark dataset (Section \ref{annotation_dataset}) for \textsc{QF-CES} evaluation.
\subsection{Prompt Design Consideration} 
Our \textsc{QF-CES-PROMPT} design comprises three key components: (\textit{1}) \textbf{Generation Prompt:} providing step-by-step instructions for comprehensive, query-relevant tabular comparisons and a final verdict summaries; (\textit{2}) \textbf{Evaluation Prompts:} for each dimension, featuring detailed criteria and scoring guidelines $(1-5)$,  that require explanations to enhance LLM response quality; and (\textit{3}) \textbf{System Message:} defining the LLM's role as a dimension-specific expert. This structured approach ensures high-quality, impartial assessments, improving the quality and relevance of comparative summaries across all dimensions.  

\subsection{Scoring Function}\label{scoring_func} 
\citet{liuug} proposed a weighted average approach to address discrete LLM scoring limitations. The final score is computed as:
\begin{align} 
    o = \sum_{k=1}^{j} p(s_{k})\times s_{k} 
\end{align}
where $s_{k}$ are possible scores and $p(s_{k})$ their LLM-determined probabilities. $p(s_{k})$ is estimated by sampling $n$ outputs $({\tt n} \approx 100)$  per input, effectively reducing scoring to a mean calculation. This method aims to enhance scoring nuance and reliability.

\subsection{Evaluation Approach}\label{appendix_correlation}
For each query $q_{i}$ in dataset $\mathcal{D}$, $i \in \{1,...,\mathcal{Q}\}$, we have $\mathcal{N}$ \textsc{QF-CES} from different models. Let $s_{ij}$ denote the $j^{th}$ \textsc{QF-CES} for query $q_{i}$, $\mathcal{M}_m$ denote the $m^{th}$ evaluation metric and $\mathcal{K}$ denote the correlation measure. \citet{bhandari-etal-2020-evaluating} defines the summary-level correlation as:
\begin{align}
    \mathcal{R}(a,b) = \frac{1}{\mathcal{Q}} \sum_{i} &\mathcal{K}([\mathcal{M}_{a}(s_{i1}),...,\mathcal{M}_{a}(s_{i\mathcal{N}})], \nonumber \\
    &[\mathcal{M}_{b}(s_{i1}),...,\mathcal{M}_{b}(s_{i\mathcal{N}})])
\end{align}
Where: $\mathcal{Q}$ is the total number of queries 
$s_{ij}$ is the \textsc{QF-CES} generated for query $q_{i}$ by model $j$
$\mathcal{M}_a$ and $\mathcal{M}_b$ are two different evaluation metrics.

\section{Dataset} \label{dataset}
We present two datasets: (\textit{1}) \textbf{\textsc{MS-Q2P}}, a novel proprietary dataset. 
(\textit{2}) \textbf{\textsc{CES-EVAL}}, a benchmark dataset for evaluating the \textsc{QF-CES} on $5$ dimensions. In this section, we discuss the dataset used, \textsc{QF-CES} evaluation metrics, annotation details, and its analysis.

\subsection{MS-Q2P Dataset }\label{q2p_dataset}
\textsc{MS-Q2P\footnote{The \textsc{MS-Q2P}, a comprehensive dataset, was provided by an e-commerce company. Details withheld for anonymity during review}} (Multi-Source Query-2-Product) comprises of a user query and the top-$k$ $(k=3)$ recommended products. Each product entry includes diverse attributes: title, description, key features, specifications, reviews, average rating, and pricing details. \textsc{MS-Q2P} consists of products from various domains like electronics, home \& kitchen, sports, clothing, shoes \& jewelry etc.  Detailed statistics of \textsc{MS-Q2P} can be found in Table \ref{tab:dataset_stats}.
\begin{table}[htp]
    \centering
    \resizebox{1\columnwidth}{!}{%
    \begin{tabular}{lcc}
         \toprule
        \textbf{Statistic} & \textbf{Value} \\
        \midrule
        \# of unique queries & 7752 \\
        Total \# of products & 23256 \\
        Average \# of reviews per product & 10 \\
        Average length of specifications per product (words) & 242.6 \\
        Average length of reviews per product (words) & 17.99 \\
        Average length of description per product (words) & 105.79 \\
        Average length of key features per product (words) & 24.64 \\
        \bottomrule
    \end{tabular}%
    }
    \caption{\textsc{MS-Q2P} dataset statistics.}
    \label{tab:dataset_stats}
\end{table}%

\subsection{CES-EVAL Dataset}\label{annotation_dataset}
We developed the \textsc{CES-EVAL} benchmark dataset to evaluate \textsc{QF-CES} across $5$ dimensions (detailed in Appendix \ref{cex_metrics}). The dataset comprises annotations for $10$ model-generated summaries per product for $50$ products from \textsc{MS-Q2P}, totaling $\mathbf{7,500}$ ratings ($3$ raters × $50$ instances × $10$ summaries × $5$ dimensions). Summaries were evaluated on a $5$-point Likert scale by three experienced students (Master's, Pre-Doctoral, Doctoral) with expertise in opinion summarization. This choice of expert raters over crowd workers was based on findings from \citet{gillick-liu-2010-non} and \citet{fabbri-etal-2021-summeval}. We employed a two-round annotation process, with discrepancies of $2$ or more points re-evaluated through discussion. Raters (male, aged $24-32$) were given comprehensive guidelines and model identities were concealed to prevent bias. Raters were compensated commensurate with their contributions to the task. Inter-rater correlations are reported in Table \ref{tab:rater_correlation}.

\subsection{Annotation Analysis} \label{anno_analysis} We evaluated the inter-rater agreement for the $3$ raters using Krippendorff's alpha coefficient ($\alpha$) \citep{Krippendorff2011ComputingKA}. For Round-I, we found the coefficient to be $0.50$ indicating \textit{moderate aggrement} ($0.41\le\alpha\le 0.60$). For Round-II, the coefficient increased to $0.80$, indicating \textit{substantial agreement} ($0.61 \le\alpha\le 0.80$). Table \ref{tab:krip_alpha} reports the dimension-wise agreement scores for both rounds. Dimension-wise analysis revealed consistently higher agreement for \foa: and \fa: consistently scoring higher in both rounds, likely due to the clear identification criteria based on format adherence. Post Round-II, \qr: and \ifo: show the most disagreement between raters, indicating challenges in consistent assessment. 

\begin{table}[htp]
    \centering
    \resizebox{1\columnwidth}{!}{%
    \begin{tabular}{lcc}
         \toprule
        & \textbf{Round-I} $\uparrow$ & \textbf{Round-II} $\uparrow$\\
        \midrule
        \cl & $0.55$ & $0.78$ \\
        \fa & $0.57$ & $0.81$\\
        \ifo & $0.44$ & $0.79$ \\
        \foa & $0.55$ & $0.82$ \\
        \qr & $0.38$ & $0.81$ \\
        \midrule
        \textbf{\textsc{AVG}} & $0.50$ & $0.80$ \\
        \bottomrule
    \end{tabular}
    }
    \caption{\textbf{Krippendorff's alpha coefficient} (${\alpha}$) for Round-I and Round-II on $5$ dimensions. As expected, Round-II shows an improvement in (${\alpha}$) scores.}
    \label{tab:krip_alpha}
\end{table}
\begin{figure}[htp]
    \centering
    \includegraphics[width=1\columnwidth]{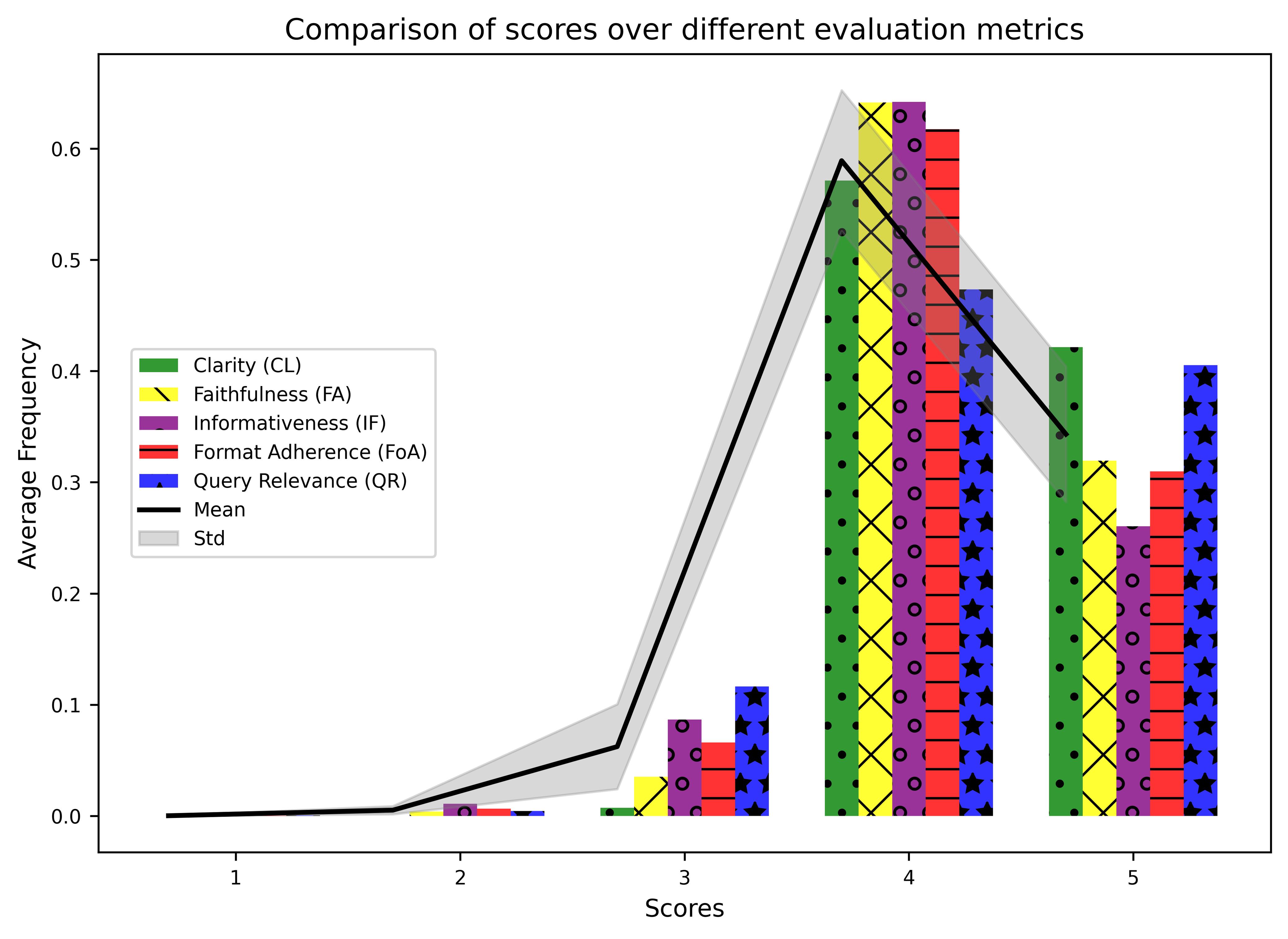}
    \caption{\textbf{Ratings Distribution.} We plot the average frequency of scores obtained by human raters across $5$ dimensions. A score of $4$ or $5$ is mostly preferred.} 
    \label{fig:ratings_plot}
\end{figure}
The average frequency of scores given by human raters across $5$ dimensions is shown in Figure \ref{fig:ratings_plot}. 

\begin{figure}[t]
    \centering
    \includegraphics[width=1.0\columnwidth]{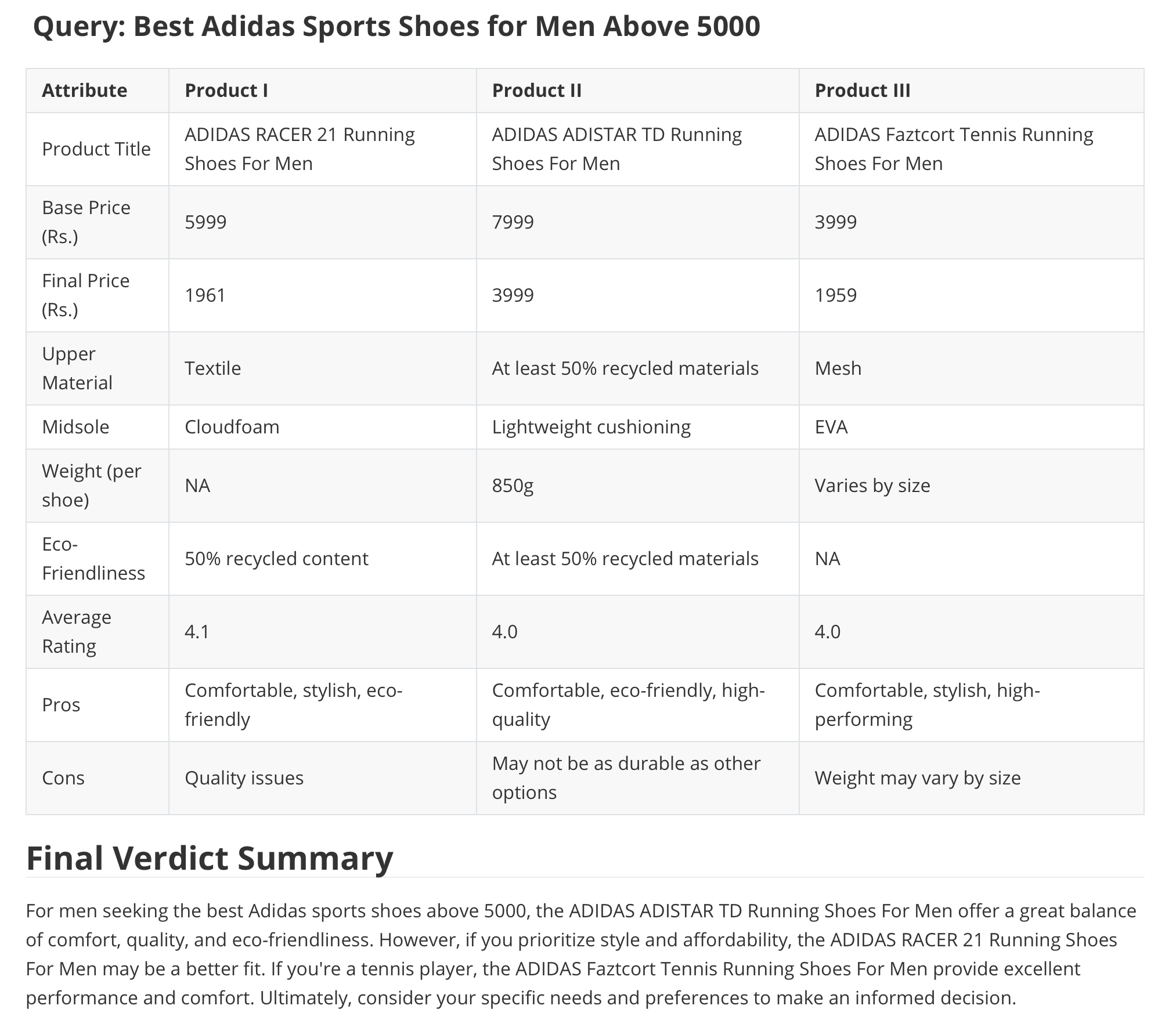}
    \caption{QF-CES generated by \qweninstruct{2}} 
    \label{fig:ex1}
\end{figure}

\section{Experiments} \label{exp}
We present the generation and evaluation of \textsc{M-OS} for \textsc{QF-CES}, using LLMs as baseline metrics, followed by implementation details.

\subsection{M-OS Models}\label{mos_models}
We use a custom prompt for the LLMs to generate \textsc{M-OS}. These models were not fine-tuned specifically for multi-source opinion summarization. We use the HuggingFace library \cite{hf} to access these $6$ open-source models: {\llamainstruct{3.1}{8}} \cite{l3}, {\mistralinstructv{7}{0.2}} \cite{m2}, {\mistralinstructv{7}{0.3}} \cite{m2}, {\gemmaone} \cite{gemma} , {\vicunavonefive} \cite{v},  {\zephyrbeta} \cite{z}   

\subsection{M-OS Evaluation}\label{mos_eval}
For evaluation of \textsc{M-OS}, we used {\llamainstruct{3.1}{70}} model \cite{l3} as our evaluator model for these reasons:  (\textit{a}) it has have outperformed {\gptturbo} and {\gptfour} on \textit{IFEval} Benchmark \cite{l3} (\textit{b}) it is ranked best amongst the open-source models on the {\tt lmsys/chatbot-arena-leaderboard}, (\textit{c}) we found its instruction following-ness to be better than alternatives, (\textit{d}) its 70B size ensures easy replication compared to {\llamainstruct{3.1}{405}}. 

\begin{table*}[t]
    \centering
    \resizebox{2\columnwidth}{!}{%
    \begin{tabular}{clcccccccccccccc}
    \toprule
           & \textbf{Evaluator LLM} & \multicolumn{2}{c}{\textbf{CL} $\uparrow$}  & \multicolumn{2}{c}{\textbf{FA} $\uparrow$} & \multicolumn{2}{c}{\textbf{IF} $\uparrow$} & \multicolumn{2}{c}{\textbf{FoA} $\uparrow$} & \multicolumn{2}{c}{\textbf{QR} $\uparrow$}\\
         \cmidrule(lr){3-4} \cmidrule(lr){5-6} \cmidrule(lr){7-8} \cmidrule(lr){9-10} \cmidrule(lr){11-12}
         & & $\rho$ & $\tau$ & $\rho$ & $\tau$ & $\rho$ & $\tau$ & $\rho$ & $\tau$ & $\rho$ & $\tau$ \\  
    \midrule
    \multirow{5}{*}{\rot{\textbf{\textsc{MS-Q2P}}}} 
       & \llamainstruct{3.1}{8} & $0.60$ & $0.50$ & $0.60$ & $0.43$ & $\underline{0.68}^*$ & $0.54$ & $0.58^*$ & $0.39$ & $\underline{0.67}$ & $\mathbf{0.59}^*$ \\
       & \mistralinstructv{7}{0.2} & $0.67$ & $0.50$ & $0.68^*$ & $\underline{0.57}^*$ & $\underline{0.68}^*$ & $\underline{0.57}^*$ & $\underline{0.69}^*$ & $\mathbf{0.54}$ & $\underline{0.67}$ & $\underline{0.55}^*$
       \\
       & \mistralinstructv{7}{0.3} & $0.68^*$ & $0.50^*$ & $0.61$ & $0.46^*$ & $0.67$ & $0.48^*$ & $0.67$ & $\underline{0.50}^*$ & $\underline{0.67}$ & $\underline{0.55}$ 
       \\
       & \llamainstruct{3.1}{70} & $\underline{0.70}^*$ & $\underline{0.56}^*$ & $\mathbf{0.77}^*$ & $\mathbf{0.63}^*$ & $\mathbf{0.82}^*$ & $\mathbf{0.65}^*$ & $\mathbf{0.73}^*$ & $\mathbf{0.54}$ & $\mathbf{0.68}^*$ & $0.46$ 
       \\
       & \gptfour & $\mathbf{0.77}^*$ & $\mathbf{0.61}^*$ & $\underline{0.75}^*$ & $\mathbf{0.63}^*$ & $0.67$ & $\underline{0.57}^*$ & $0.68^*$ & $\underline{0.50}^*$ & $\mathbf{0.68}^*$ & $\mathbf{0.59}^*$ \\
    \bottomrule
    \end{tabular}
    }
    \caption{\textit{Spearman} ($\rho$) and \textit{Kendall Tau} ($\tau$) correlations at summary-level on $5$ dimensions: {\cl} (CL), {\fa} (FA), {\ifo} (IF), {\foa} (FoA) and {\qr} (QR) for the \textsc{MS-Q2P} dataset. \textsc{Llama-3.1-70B-Instruct} demonstrates the highest correlations across most dimensions, indicating strong agreement with human evaluations, followed by \textsc{GPT-4o}.  Best performing values are boldfaced, and the second best underlined. $*$ represents significant performance (p-value $< 0.05$) Refer Figure \ref{fig:different_evaluators} for graphical representation of model-wise performance across different evaluators.}
\label{tab:main_results_table}
\end{table*}

\begin{table}[t]
    \centering
    \resizebox{1\columnwidth}{!}{%
    \begin{tabular}{lcccccc}
    \toprule
         \textbf{LLM} & \textbf{CL} $\uparrow$ & \textbf{FA} $\uparrow$ & \textbf{IF} $\uparrow$ & \textbf{FoA} $\uparrow$ & \textbf{QR} $\uparrow$ & \textbf{Average} $\uparrow$ \\

    \midrule
        \gemmaone & $4.35$ & $4.39$ & $3.95$ & $3.58$ & $3.79$ & $4.01$ \\
        \llamainstruct{3.1}{8} & $4.54$ & $4.25$ & $4.17$ & $4.39$ & $4.24$ & $4.32$ \\
        \mistralinstructv{7}{0.2} & $\underline{4.60}$ & $4.24$ & $4.06$ & $4.22$ & $4.45$ & $4.31$ \\
        \mistralinstructv{7}{0.3} & $4.28$ & $4.28$ & $4.19$ & $4.25$ & $4.43$ & $4.29$ \\
        \qweninstruct{2} & $4.47$ & $\underline{4.41}$ & $\mathbf{4.58}$ & $4.36$ & $\mathbf{4.63}$ & $\underline{4.49}$ \\
        \gemmatwo & $4.19$ & $4.00$ & $4.19$ & $4.17$ & $4.37$ & $4.18$ \\
        \mixtral & $4.35$ & $4.29$ & $4.17$ & $4.35$ & $4.43$ & $4.32$ \\
        \llamainstruct{3.1}{70} & $\underline{4.55}$ & $4.36$ & $4.47$ & $\underline{4.41}$ & $4.03$ & $4.36$ \\
        \gptfourstandard & $\mathbf{4.81}$ & $\mathbf{4.53}$ & $\underline{4.50}$ & $\mathbf{4.61}$ & $4.32$ & $\mathbf{4.55}$ \\
        \qweninstruct{2}-DIA & $4.30$ & $4.33$ & $3.81$ & $4.18$ & $3.89$ & $4.10$ \\
    \bottomrule
    \end{tabular}
    }
    \caption{Model-wise averaged annotator ratings of \textsc{QF-CES} along $5$ dimensions: {\cl} (CL), {\fa} (FA), {\ifo} (IF), {\foa} (FoA) and {\qr} (QR) Best scores are in \textbf{bold}, second-best are \underline{underlined}. \qweninstruct{2}-DIA\ represents inputting raw data directly to LLM. Refer Figure \ref{fig:cex_human} for graphical representation.}
    \label{tab:human_cex_score_round2}
\end{table}

\begin{table}[t]
    \centering
    \resizebox{1\columnwidth}{!}{%
    \begin{tabular}{lccccccc}
    \toprule
         \textbf{Model} & \textbf{FL} $\uparrow$ & \textbf{CO} $\uparrow$ & \textbf{AC} $\uparrow$ & \textbf{FF} $\uparrow$ & \textbf{RL} $\uparrow$ & \textbf{SC} $\uparrow$ & \textbf{SP} $\uparrow$\\
    \midrule
        \mistralinstructv{7}{0.3} & $4.73$ & $4.64$ & $4.09$ & $4.08$ & $4.05$ & $4.24$ & $4.06$ \\
       \llamainstruct{3.1}{8} & $4.90$ & $4.63$ & $3.94$ & $4.05$ & $4.01$ & $4.16$ & $3.63$ \\
        \mistralinstructv{7}{0.2} & $4.70$ & $4.52$ & $3.94$ & $4.05$ & $4.01$ & $4.13$ & $3.99$ \\
        \gemmaone & $4.30$ & $4.35$ & $3.82$ & $4.04$ & $3.99$ & $3.97$ & $3.25$ \\
        \vicunavonefive & $3.89$ & $3.76$ & $3.51$ & $3.92$ & $3.62$ & $3.50$ & $3.06$ \\
        \zephyrbeta  & $4.79$ & $4.56$ & $3.86$ & $4.14$ & $4.06$ & $4.09$ & $3.79$ \\
    \bottomrule
    \end{tabular}
    }
    \caption{\textsc{M-OS} Model-wise performance across $7$ dimensions: {\fl} (FL), {\coh} (CO), {\asp} (AC), {\fa} (FF), {\rel} (RL), {\sent} (SC), and {\spec} (SP), evaluated by {\llamainstruct{3.1}{70}} over $n=100$ generations. Refer Figure \ref{fig:opsum} for graphical representation.}
    \label{tab:opsum}
\end{table}

\subsection{QF-CES Models}\label{ces_models}
Baselines: In our experiments, we adopt a range of recent widely-used LLMs. For close-sourced LLM (accessible through APIs), we evaluate OpenAI’s {\gptfourstandard} \cite{openai}. For open-source LLMs, use the HuggingFace library \cite{hf} to access these models and experimented with {\llamainstruct{3.1}{70}} \cite{l3}, {\llamainstruct{3.1}{8}} \cite{l3},{\gemmaone} \cite{gemma}, {\gemmatwo} \cite{gemma}, {\mistralinstructv{7}{0.2}} \citet{m2}, {\mistralinstructv{7}{0.3}} \cite{m2}, {\qweninstructnew} \cite{q} and {\mixtralnew} \cite{mix} as baselines. 

\subsection{QF-CES Evaluation}\label{ces_eval}
For evaluation of \textsc{QF-CES}, we used one proprietary LLM which is OpenAI’s {\gptfour} \cite{openai}  and $4$ open source LLMs as evaluators: {\llamainstruct{3.1}{70}} \cite{l3}, {\llamainstruct{3.1}{8}} \cite{l3}, {\mistralinstructv{7}{0.2}} \cite{m2} and {\mistralinstructv{7}{0.3}} \cite{m2}.
These models were chosen based on their performance on benchmarks, ranking on {\tt lmsys/chatbot-arena-leaderboard}, instruction-following capabilities, replicability, and popularity on Hugging Face.

\subsection{Implementation Details}\label{implement}
 In \textbf{summary generation} (\textsc{M-OS} \& \textsc{QF-CES}), we configure open-source LLMs with \verb|top_k=25, top_p=0.95, number of beams=3| and \verb|temperature=0.2| to produce deterministic outputs capturing product technicalities. For OpenAI's {\gptfourstandard} \citep{openai}, we set \verb|decoding temperature=0| for increased determinism. 
 
 In \textbf{summary evaluation}, open-source LLMs use \verb|n=100, temperature=0.2| to account for stochasticity while maintaining consistency. {\gptfour} \citep{openai} again uses \verb|decoding temperature=0|. 
 
 All experiments run on 8 NVIDIA A100-SXM4-80GB clusters, ensuring robust computational capacity for our analyses.

\section{Results and Analysis} \label{results_analysis}
Our obtained results are provided in Tables \ref{tab:main_results_table}, \ref{tab:human_cex_score_round2} and \ref{tab:opsum}.

\textbf{\textsc{M-OS} Results:} Table \ref{tab:opsum} presents averaged scores assigned by {\llamainstruct{3.1}{70}} \cite{l3} across $7$ dimensions for each model evaluated for \textsc{M-OS} generation, with Figure \ref{fig:opsum} providing a graphical representation of these results.  {\mistralinstructv{7}{0.3}} \cite{m2} and was selected to generate \textsc{M-OS} for use in \textsc{QF-CES}.

\textbf{\textsc{QF-CES} Results:}  Table \ref{tab:human_cex_score_round2} (Refer to Figure 5 for a graphical view) presents averaged annotator ratings across $5$ dimensions for each evaluated model, with Figure \ref{fig:cex_human} providing a graphical representation. {\gptfour} \citep{openai} excelled, leading in 4 dimensions and competing strongly in {\qr}. Among open-source models, {\qweninstruct{2}} \citep{q} achieved the highest average score, particularly in {\ifo} and {\qr}. While larger models like {\gptfour} and {\llamainstruct{3.1}{70}} \citep{l3} generally performed better, smaller models like {\qweninstruct{2}} \cite{q} showed competitive results in specific areas, highlighting the importance of task-specific model selection. {\qweninstruct{2}}'s \cite{q} performance using a \textsc{Direct Input Approach} was inferior, especially in {\ifo} and {\qr}, likely due to data overload when generating \textsc{QF-CES} for top-$3$ products. These findings validate our \textsc{M-OS} approach as an effective intermediate step for high-quality \textsc{QF-CES} generation.

\textbf{\textsc{QF-CES} Evaluation Results:} Table \ref{tab:main_results_table} report the summary-level correlation scores on the \textsc{CES-EVAL} dataset, for $4$ open-source models and one closed-source model. Overall, {\llamainstruct{3.1}{70}} ranks as the best evaluator achieving  $\mathbf{0.74}$ average spearman correlation, outperforming OpenAI’s {\gptfour} \cite{openai} across all dimensions. 
Figure \ref{fig:different_evaluators} provides a visual representation of the scores given by different LLMs acting as evaluators for \verb|n=100| generations for the $5$ dimensions of \textsc{QF-CES} across various models. This multi-model evaluation approach offers a comprehensive view of model performance, reinforcing the reliability of our findings.

\textbf{Time Efficiency Results:}\label{time} To account for the stochasticity of LLMs, we generated $50$ iterations of each \textsc{QF-CES} (\verb|n=50|), which also helps mitigate instances where an LLM might loop until reaching the maximum token length, potentially skewing time measurements.  We recorded generation times for both \textsc{M-OS} and \textsc{DIA} methods, excluding \textsc{M-OS} generation time to reflect real-world pre-computation scenarios.
Analysis of average generation time per query (Figure \ref{fig:time}) shows \textsc{M-OS} significantly accelerates inference by $\mathbf{40\%}$ compared to \textsc{DIA}. Queries $24$ and $27$, involving complex electronics specifications, demonstrated \textsc{M-OS}'s efficiency in condensing large data volumes, outpacing \textsc{DIA}'s raw data processing.
\begin{figure}[htp]
    \centering
    \includegraphics[width=1\columnwidth]{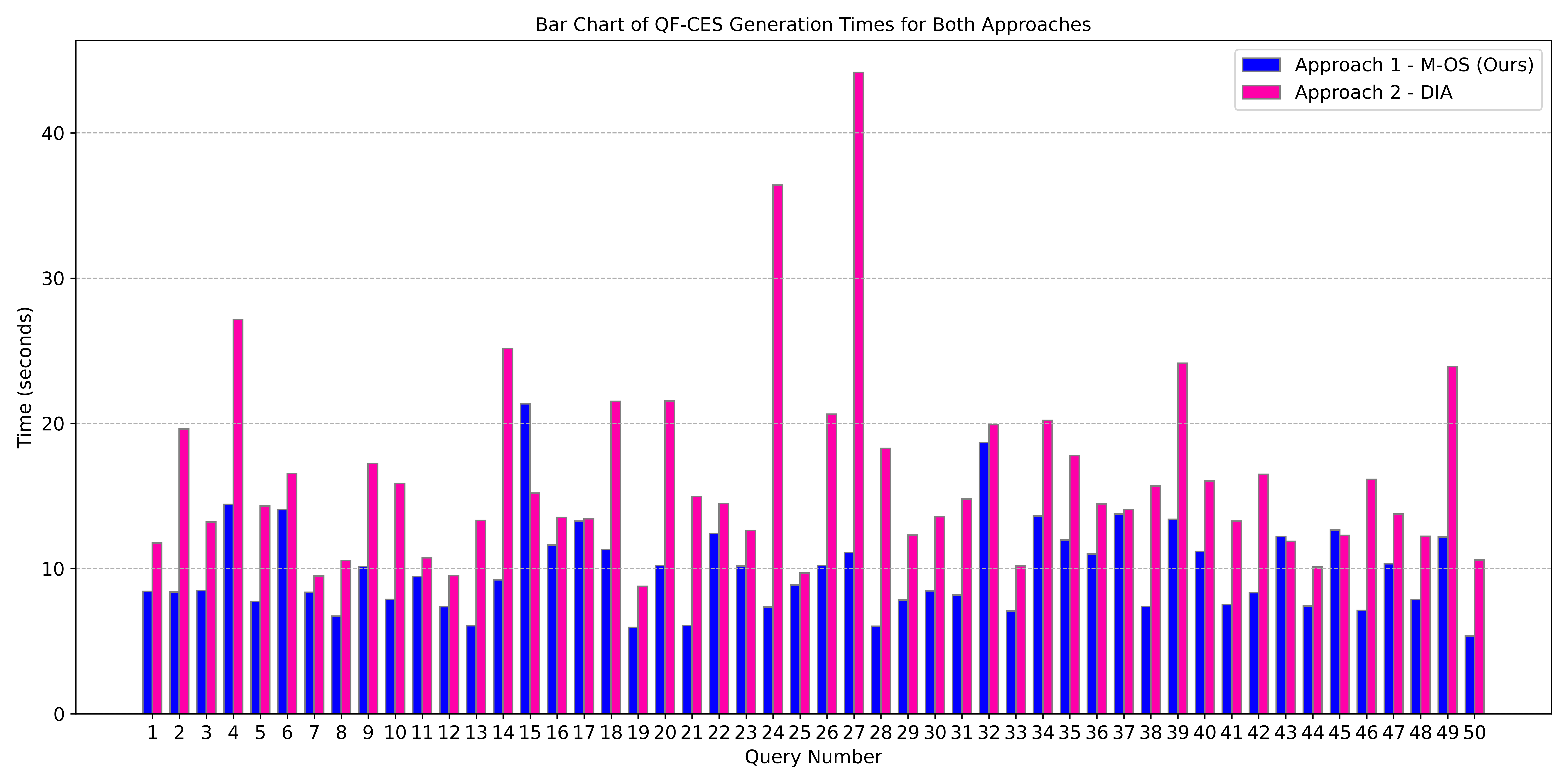}
    \caption{Comparison of inference times for \textsc{QF-CES} generation using \textsc{M-OS} and \textsc{DIA} approach. Each data point represents the average of $50$ generations per query.} 
    \label{fig:time}
\end{figure}

\section{Conclusion \& Future Work}
This paper introduces Query-Focused Comparative Explainable Summarization \textsc{QF-CES}, a novel task that addresses the limitations of traditional opinion summarization in e-commerce recommendation systems. By leveraging LLMs and Multi-Source Opinion Summarization \textsc{M-OS}, we present a comprehensive approach to generate query-specific, comparative summaries of recommended products. The framework, validated through the \textsc{MS-Q2P} dataset and extensive evaluations, showed {\gptfourstandard} superior performance, with {\qweninstruct{2}} as a strong open-source contender for \textsc{QF-CES}. The evaluation using \textsc{QF-CES-PROMPT} with {\llamainstruct{3.1}{70}} yielded an average \textit{Spearman correlation} of $\mathbf{0.74}$ with human judgments, highlighting its reliability. Future work will focus on improving LLM performance on complex products with extensive specifications, reducing inference latency while maintaining high summary quality, refining prompting strategies to better select relevant attributes based on user queries, and exploring the applicability of \textsc{QF-CES-PROMPT} to other domains beyond e-commerce to assess its generalizability and necessary adaptations.

\section*{Limitations}
\begin{enumerate}
    \item We evaluate using {\gptfour} for \textsc{QF-CES}, omitting {\gptfourstandard} due to cost constraints. Our primary goal was to design prompts applicable to open-source and closed-source models for generating and evaluating \textsc{M-OS} and \textsc{QF-CES}.
    \item Our \textsc{QF-CES-PROMPT} targets a specific dimension of comparative summarization. Its broader applicability requires further study and potential prompt adjustments.
    \item The \textsc{M2-Q2P} dataset's limitation of only $10$ reviews highlights a broader issue in opinion summarization. Future benchmarks should incorporate datasets with more reviews.
`   \item During \textsc{QF-CES} generation, LLMs occasionally struggled with products having 
     extensive specifications, resulting in incomplete or stalled summaries.
    \item\textsc{QF-CES} with \textsc{M-OS} consistently outperformed \textsc{DIA} in inference latency across $50$ queries. However, a larger query set is necessary to solidify and generalize these findings.
\end{enumerate}

\section*{Ethical Considerations}
We engaged $3$ raters with diverse academic backgrounds: a Master's student, a Pre-Doctoral researcher, and a Doctoral candidate. All were male, aged 24-32, with publications or active research in opinion summarization. Raters were compensated appropriately for their contributions.

\textsc{QF-CES-PROMPTS} generate and evaluate \textsc{QF-CES} across $5$ dimensions, aiding the assessment of NLG-produced comparative summaries. While insightful, these prompts may occasionally produce hallucinations, especially in complex cases. We urge judicious use and validation of reliability for specific applications. Researchers should verify prompt appropriateness before integration into evaluation processes, ensuring careful application in real-world scenarios.

\bibliography{anthology,custom}

\begin{thebibliography}{44}
\expandafter\ifx\csname natexlab\endcsname\relax\def\natexlab#1{#1}\fi

\bibitem[{AI@Meta(2024)}]{l3}
AI@Meta. 2024.
\newblock \href {https://github.com/meta-llama/llama3/blob/main/MODEL_CARD.md} {Llama 3 model card}.

\bibitem[{Bhandari et~al.(2020)Bhandari, Gour, Ashfaq, Liu, and Neubig}]{bhandari-etal-2020-evaluating}
Manik Bhandari, Pranav~Narayan Gour, Atabak Ashfaq, Pengfei Liu, and Graham Neubig. 2020.
\newblock \href {https://doi.org/10.18653/v1/2020.emnlp-main.751} {Re-evaluating evaluation in text summarization}.
\newblock In \emph{Proceedings of the 2020 Conference on Empirical Methods in Natural Language Processing (EMNLP)}, pages 9347--9359, Online. Association for Computational Linguistics.

\bibitem[{Chen et~al.(2018)Chen, Zhang, Liu, and Ma}]{chen2018neural}
Chong Chen, Min Zhang, Yiqun Liu, and Shaoping Ma. 2018.
\newblock \href {https://doi.org/10.1145/3178876.3186070} {Neural attentional rating regression with review-level explanations}.
\newblock In \emph{Proceedings of the 2018 World Wide Web Conference}, WWW '18, page 1583–1592, Republic and Canton of Geneva, CHE. International World Wide Web Conferences Steering Committee.

\bibitem[{Chiang and Lee(2023{\natexlab{a}})}]{chiang-lee-2023-large}
Cheng-Han Chiang and Hung-yi Lee. 2023{\natexlab{a}}.
\newblock \href {https://doi.org/10.18653/v1/2023.acl-long.870} {Can large language models be an alternative to human evaluations?}
\newblock In \emph{Proceedings of the 61st Annual Meeting of the Association for Computational Linguistics (Volume 1: Long Papers)}, pages 15607--15631, Toronto, Canada. Association for Computational Linguistics.

\bibitem[{Chiang and Lee(2023{\natexlab{b}})}]{chianggg}
Cheng-Han Chiang and Hung-yi Lee. 2023{\natexlab{b}}.
\newblock \href {https://doi.org/10.18653/v1/2023.acl-long.870} {Can large language models be an alternative to human evaluations?}
\newblock In \emph{Proceedings of the 61st Annual Meeting of the Association for Computational Linguistics (Volume 1: Long Papers)}, pages 15607--15631, Toronto, Canada. Association for Computational Linguistics.

\bibitem[{Chiang and Lee(2023{\natexlab{c}})}]{closer}
Cheng-Han Chiang and Hung-yi Lee. 2023{\natexlab{c}}.
\newblock \href {https://doi.org/10.18653/v1/2023.findings-emnlp.599} {A closer look into using large language models for automatic evaluation}.
\newblock In \emph{Findings of the Association for Computational Linguistics: EMNLP 2023}, pages 8928--8942, Singapore. Association for Computational Linguistics.

\bibitem[{Chiang et~al.(2023)Chiang, Li, Lin, Sheng, Wu, Zhang, Zheng, Zhuang, Zhuang, Gonzalez, Stoica, and Xing}]{v}
Wei-Lin Chiang, Zhuohan Li, Zi~Lin, Ying Sheng, Zhanghao Wu, Hao Zhang, Lianmin Zheng, Siyuan Zhuang, Yonghao Zhuang, Joseph~E. Gonzalez, Ion Stoica, and Eric~P. Xing. 2023.
\newblock \href {https://lmsys.org/blog/2023-03-30-vicuna/} {Vicuna: An open-source chatbot impressing gpt-4 with 90\%* chatgpt quality}.

\bibitem[{Colas et~al.(2023)Colas, Araki, Zhou, Wang, and Feng}]{colas2023knowrec}
Anthony Colas, Jun Araki, Zhengyu Zhou, Bingqing Wang, and Zhe Feng. 2023.
\newblock \href {https://doi.org/10.18653/v1/2023.blackboxnlp-1.1} {Knowledge-grounded natural language recommendation explanation}.
\newblock In \emph{Proceedings of the 6th BlackboxNLP Workshop: Analyzing and Interpreting Neural Networks for NLP}, pages 1--15, Singapore. Association for Computational Linguistics.

\bibitem[{Echterhoff et~al.(2023)Echterhoff, Yan, and McAuley}]{echterhoff2023}
Jessica~Maria Echterhoff, An~Yan, and Julian McAuley. 2023.
\newblock \href {https://api.semanticscholar.org/CorpusID:259375808} {Comparing apples to apples: Generating aspect-aware comparative sentences from user reviews}.
\newblock \emph{ArXiv}, abs/2307.03691.

\bibitem[{Fabbri et~al.(2021)Fabbri, Kry{\'s}ci{\'n}ski, McCann, Xiong, Socher, and Radev}]{fabbri-etal-2021-summeval}
Alexander~R. Fabbri, Wojciech Kry{\'s}ci{\'n}ski, Bryan McCann, Caiming Xiong, Richard Socher, and Dragomir Radev. 2021.
\newblock \href {https://doi.org/10.1162/tacl_a_00373} {{S}umm{E}val: Re-evaluating summarization evaluation}.
\newblock \emph{Transactions of the Association for Computational Linguistics}, 9:391--409.

\bibitem[{Fu et~al.(2023)Fu, Ng, Jiang, and Liu}]{fu}
Jinlan Fu, See-Kiong Ng, Zhengbao Jiang, and Pengfei Liu. 2023.
\newblock \href {http://arxiv.org/abs/2302.04166} {Gptscore: Evaluate as you desire}.

\bibitem[{Gao et~al.(2024)Gao, Wang, Fang, Chen, Han, and Shang}]{dre2023}
Shen Gao, Yifan Wang, Jiabao Fang, Lisi Chen, Peng Han, and Shuo Shang. 2024.
\newblock \href {http://arxiv.org/abs/2404.06311} {Dre: Generating recommendation explanations by aligning large language models at data-level}.

\bibitem[{Gillick and Liu(2010)}]{gillick-liu-2010-non}
Dan Gillick and Yang Liu. 2010.
\newblock \href {https://aclanthology.org/W10-0722} {Non-expert evaluation of summarization systems is risky}.
\newblock In \emph{Proceedings of the {NAACL} {HLT} 2010 Workshop on Creating Speech and Language Data with {A}mazon{'}s Mechanical Turk}, pages 148--151, Los Angeles. Association for Computational Linguistics.

\bibitem[{Im et~al.(2021)Im, Kim, Lee, Cho, and Chung}]{im}
Jinbae Im, Moonki Kim, Hoyeop Lee, Hyunsouk Cho, and Sehee Chung. 2021.
\newblock \href {https://doi.org/10.18653/v1/2021.acl-long.33} {Self-supervised multimodal opinion summarization}.
\newblock In \emph{Proceedings of the 59th Annual Meeting of the Association for Computational Linguistics and the 11th International Joint Conference on Natural Language Processing (Volume 1: Long Papers)}, pages 388--403, Online. Association for Computational Linguistics.

\bibitem[{Iso et~al.(2022)Iso, Wang, Angelidis, and Suhara}]{collaborative2022}
Hayate Iso, Xiaolan Wang, Stefanos Angelidis, and Yoshihiko Suhara. 2022.
\newblock {C}omparative {O}pinion {S}ummarization via {C}ollaborative {D}ecoding.
\newblock In \emph{Findings of the Association for Computational Linguistics (ACL)}.

\bibitem[{Jiang et~al.(2023)Jiang, Sablayrolles, Mensch, Bamford, Chaplot, de~las Casas, Bressand, Lengyel, Lample, Saulnier, Lavaud, Lachaux, Stock, Scao, Lavril, Wang, Lacroix, and Sayed}]{m2}
Albert~Q. Jiang, Alexandre Sablayrolles, Arthur Mensch, Chris Bamford, Devendra~Singh Chaplot, Diego de~las Casas, Florian Bressand, Gianna Lengyel, Guillaume Lample, Lucile Saulnier, Lélio~Renard Lavaud, Marie-Anne Lachaux, Pierre Stock, Teven~Le Scao, Thibaut Lavril, Thomas Wang, Timothée Lacroix, and William~El Sayed. 2023.
\newblock \href {http://arxiv.org/abs/2310.06825} {Mistral 7b}.

\bibitem[{Jiang et~al.(2024)Jiang, Sablayrolles, Roux, Mensch, Savary, Bamford, Chaplot, de~las Casas, Hanna, Bressand, Lengyel, Bour, Lample, Lavaud, Saulnier, Lachaux, Stock, Subramanian, Yang, Antoniak, Scao, Gervet, Lavril, Wang, Lacroix, and Sayed}]{mix}
Albert~Q. Jiang, Alexandre Sablayrolles, Antoine Roux, Arthur Mensch, Blanche Savary, Chris Bamford, Devendra~Singh Chaplot, Diego de~las Casas, Emma~Bou Hanna, Florian Bressand, Gianna Lengyel, Guillaume Bour, Guillaume Lample, Lélio~Renard Lavaud, Lucile Saulnier, Marie-Anne Lachaux, Pierre Stock, Sandeep Subramanian, Sophia Yang, Szymon Antoniak, Teven~Le Scao, Théophile Gervet, Thibaut Lavril, Thomas Wang, Timothée Lacroix, and William~El Sayed. 2024.
\newblock \href {http://arxiv.org/abs/2401.04088} {Mixtral of experts}.

\bibitem[{Kocmi and Federmann(2023)}]{kocmi-federmann-2023-large}
Tom Kocmi and Christian Federmann. 2023.
\newblock \href {https://aclanthology.org/2023.eamt-1.19} {Large language models are state-of-the-art evaluators of translation quality}.
\newblock In \emph{Proceedings of the 24th Annual Conference of the European Association for Machine Translation}, pages 193--203, Tampere, Finland. European Association for Machine Translation.

\bibitem[{Krippendorff(2011)}]{Krippendorff2011ComputingKA}
Klaus Krippendorff. 2011.
\newblock \href {https://api.semanticscholar.org/CorpusID:59901023} {Computing krippendorff's alpha-reliability}.

\bibitem[{Le and Lauw(2021)}]{constraints2021}
Trung-Hoang Le and Hady~W. Lauw. 2021.
\newblock \href {https://doi.org/10.1145/3437963.3441754} {Explainable recommendation with comparative constraints on product aspects}.
\newblock In \emph{Proceedings of the 14th ACM International Conference on Web Search and Data Mining}, WSDM '21, page 967–975, New York, NY, USA. Association for Computing Machinery.

\bibitem[{Li et~al.(2020{\natexlab{a}})Li, Yuan, Xu, Wu, He, and Zhou}]{LI}
Haoran Li, Peng Yuan, Song Xu, Youzheng Wu, Xiaodong He, and Bowen Zhou. 2020{\natexlab{a}}.
\newblock \href {https://doi.org/10.1609/aaai.v34i05.6332} {Aspect-aware multimodal summarization for chinese e-commerce products}.
\newblock \emph{Proceedings of the AAAI Conference on Artificial Intelligence}, 34(05):8188--8195.

\bibitem[{Li et~al.(2020{\natexlab{b}})Li, Zhang, and Chen}]{li2020generate}
Lei Li, Yongfeng Zhang, and Li~Chen. 2020{\natexlab{b}}.
\newblock \href {https://doi.org/10.1145/3340531.3411992} {Generate neural template explanations for recommendation}.
\newblock In \emph{Proceedings of the 29th ACM International Conference on Information \& Knowledge Management}, CIKM '20, page 755–764, New York, NY, USA. Association for Computing Machinery.

\bibitem[{Li et~al.(2020{\natexlab{c}})Li, Zhang, and Chen}]{LiLi}
Lei Li, Yongfeng Zhang, and Li~Chen. 2020{\natexlab{c}}.
\newblock \href {https://doi.org/10.1145/3340531.3411992} {Generate neural template explanations for recommendation}.
\newblock In \emph{Proceedings of the 29th ACM International Conference on Information \& Knowledge Management}, CIKM '20, page 755–764, New York, NY, USA. Association for Computing Machinery.

\bibitem[{Li et~al.(2021)Li, Zhang, and Chen}]{peter2023peter}
Lei Li, Yongfeng Zhang, and Li~Chen. 2021.
\newblock \href {https://doi.org/10.18653/v1/2021.acl-long.383} {Personalized transformer for explainable recommendation}.
\newblock In \emph{Proceedings of the 59th Annual Meeting of the Association for Computational Linguistics and the 11th International Joint Conference on Natural Language Processing (Volume 1: Long Papers)}, pages 4947--4957, Online. Association for Computational Linguistics.

\bibitem[{Lin(2004)}]{ro}
Chin-Yew Lin. 2004.
\newblock \href {https://aclanthology.org/W04-1013} {{ROUGE}: A package for automatic evaluation of summaries}.
\newblock In \emph{Text Summarization Branches Out}, pages 74--81, Barcelona, Spain. Association for Computational Linguistics.

\bibitem[{Liu et~al.(2023)Liu, Iter, Xu, Wang, Xu, and Zhu}]{liuug}
Yang Liu, Dan Iter, Yichong Xu, Shuohang Wang, Ruochen Xu, and Chenguang Zhu. 2023.
\newblock \href {https://doi.org/10.18653/v1/2023.emnlp-main.153} {{G}-eval: {NLG} evaluation using gpt-4 with better human alignment}.
\newblock In \emph{Proceedings of the 2023 Conference on Empirical Methods in Natural Language Processing}, pages 2511--2522, Singapore. Association for Computational Linguistics.

\bibitem[{Ni et~al.(2019)Ni, Li, and McAuley}]{Ji}
Jianmo Ni, Jiacheng Li, and Julian McAuley. 2019.
\newblock \href {https://doi.org/10.18653/v1/D19-1018} {Justifying recommendations using distantly-labeled reviews and fine-grained aspects}.
\newblock In \emph{Proceedings of the 2019 Conference on Empirical Methods in Natural Language Processing and the 9th International Joint Conference on Natural Language Processing (EMNLP-IJCNLP)}, pages 188--197, Hong Kong, China. Association for Computational Linguistics.

\bibitem[{OpenAI(2023)}]{openai}
OpenAI. 2023.
\newblock \href {https://arxiv.org/abs/2303.08774} {Gpt-4 technical report}.
\newblock \emph{ArXiv}, abs/2303.08774.

\bibitem[{Papineni et~al.(2002)Papineni, Roukos, Ward, and Zhu}]{bleu}
Kishore Papineni, Salim Roukos, Todd Ward, and Wei-Jing Zhu. 2002.
\newblock \href {https://doi.org/10.3115/1073083.1073135} {{B}leu: a method for automatic evaluation of machine translation}.
\newblock In \emph{Proceedings of the 40th Annual Meeting of the Association for Computational Linguistics}, pages 311--318, Philadelphia, Pennsylvania, USA. Association for Computational Linguistics.

\bibitem[{Peng et~al.(2024)Peng, Chen, Lin, Huang, Hu, Guo, Kong, Hu, Wu, and Wang}]{uncertainty2023}
Yicui Peng, Hao Chen, Chingsheng Lin, Guo Huang, Jinrong Hu, Hui Guo, Bin Kong, Shu Hu, Xi~Wu, and Xin Wang. 2024.
\newblock \href {http://arxiv.org/abs/2402.03366} {Uncertainty-aware explainable recommendation with large language models}.

\bibitem[{Siledar et~al.(2023)Siledar, Makwana, and Bhattacharyya}]{TJ-MOS}
Tejpalsingh Siledar, Jigar Makwana, and Pushpak Bhattacharyya. 2023.
\newblock \href {https://doi.org/10.1145/3570991.3571035} {Aspect-sentiment-based opinion summarization using multiple information sources}.
\newblock In \emph{Proceedings of the 6th Joint International Conference on Data Science \& Management of Data (10th ACM IKDD CODS and 28th COMAD)}, CODS-COMAD '23, page 55–61, New York, NY, USA. Association for Computing Machinery.

\bibitem[{Siledar et~al.(2024)Siledar, Nath, Muddu, Rangaraju, Nath, Bhattacharyya, Banerjee, Patil, Singh, Chelliah, and Garera}]{op}
Tejpalsingh Siledar, Swaroop Nath, Sankara Sri Raghava~Ravindra Muddu, Rupasai Rangaraju, Swaprava Nath, Pushpak Bhattacharyya, Suman Banerjee, Amey Patil, Sudhanshu~Shekhar Singh, Muthusamy Chelliah, and Nikesh Garera. 2024.
\newblock \href {http://arxiv.org/abs/2402.11683} {One prompt to rule them all: Llms for opinion summary evaluation}.

\bibitem[{Tan et~al.(2021)Tan, Xu, Ge, Li, Chen, and Zhang}]{count}
Juntao Tan, Shuyuan Xu, Yingqiang Ge, Yunqi Li, Xu~Chen, and Yongfeng Zhang. 2021.
\newblock \href {https://doi.org/10.1145/3459637.3482420} {Counterfactual explainable recommendation}.
\newblock In \emph{Proceedings of the 30th ACM International Conference on Information \& Knowledge Management}, CIKM '21, page 1784–1793, New York, NY, USA. Association for Computing Machinery.

\bibitem[{Tay(2019)}]{tay-2019}
Wenyi Tay. 2019.
\newblock \href {https://doi.org/10.18653/v1/P19-2005} {Not all reviews are equal: Towards addressing reviewer biases for opinion summarization}.
\newblock In \emph{Proceedings of the 57th Annual Meeting of the Association for Computational Linguistics: Student Research Workshop}, pages 34--42, Florence, Italy. Association for Computational Linguistics.

\bibitem[{Team et~al.(2024)Team, Mesnard, Hardin, Dadashi, Bhupatiraju, Pathak, Sifre, Rivière, Kale, Love, Tafti, Hussenot, Sessa, Chowdhery, Roberts, Barua, Botev, Castro-Ros, Slone, Héliou, Tacchetti, Bulanova, Paterson, Tsai, Shahriari, Lan, Choquette-Choo, Crepy, Cer, Ippolito, Reid, Buchatskaya, Ni, Noland, Yan, Tucker, Muraru, Rozhdestvenskiy, Michalewski, Tenney, Grishchenko, Austin, Keeling, Labanowski, Lespiau, Stanway, Brennan, Chen, Ferret, Chiu, Mao-Jones, Lee, Yu, Millican, Sjoesund, Lee, Dixon, Reid, Mikuła, Wirth, Sharman, Chinaev, Thain, Bachem, Chang, Wahltinez, Bailey, Michel, Yotov, Chaabouni, Comanescu, Jana, Anil, McIlroy, Liu, Mullins, Smith, Borgeaud, Girgin, Douglas, Pandya, Shakeri, De, Klimenko, Hennigan, Feinberg, Stokowiec, hui Chen, Ahmed, Gong, Warkentin, Peran, Giang, Farabet, Vinyals, Dean, Kavukcuoglu, Hassabis, Ghahramani, Eck, Barral, Pereira, Collins, Joulin, Fiedel, Senter, Andreev, and Kenealy}]{gemma}
Gemma Team, Thomas Mesnard, Cassidy Hardin, Robert Dadashi, Surya Bhupatiraju, Shreya Pathak, Laurent Sifre, Morgane Rivière, Mihir~Sanjay Kale, Juliette Love, Pouya Tafti, Léonard Hussenot, Pier~Giuseppe Sessa, Aakanksha Chowdhery, Adam Roberts, Aditya Barua, Alex Botev, Alex Castro-Ros, Ambrose Slone, Amélie Héliou, Andrea Tacchetti, Anna Bulanova, Antonia Paterson, Beth Tsai, Bobak Shahriari, Charline~Le Lan, Christopher~A. Choquette-Choo, Clément Crepy, Daniel Cer, Daphne Ippolito, David Reid, Elena Buchatskaya, Eric Ni, Eric Noland, Geng Yan, George Tucker, George-Christian Muraru, Grigory Rozhdestvenskiy, Henryk Michalewski, Ian Tenney, Ivan Grishchenko, Jacob Austin, James Keeling, Jane Labanowski, Jean-Baptiste Lespiau, Jeff Stanway, Jenny Brennan, Jeremy Chen, Johan Ferret, Justin Chiu, Justin Mao-Jones, Katherine Lee, Kathy Yu, Katie Millican, Lars~Lowe Sjoesund, Lisa Lee, Lucas Dixon, Machel Reid, Maciej Mikuła, Mateo Wirth, Michael Sharman, Nikolai Chinaev, Nithum Thain, Olivier Bachem,
  Oscar Chang, Oscar Wahltinez, Paige Bailey, Paul Michel, Petko Yotov, Rahma Chaabouni, Ramona Comanescu, Reena Jana, Rohan Anil, Ross McIlroy, Ruibo Liu, Ryan Mullins, Samuel~L Smith, Sebastian Borgeaud, Sertan Girgin, Sholto Douglas, Shree Pandya, Siamak Shakeri, Soham De, Ted Klimenko, Tom Hennigan, Vlad Feinberg, Wojciech Stokowiec, Yu~hui Chen, Zafarali Ahmed, Zhitao Gong, Tris Warkentin, Ludovic Peran, Minh Giang, Clément Farabet, Oriol Vinyals, Jeff Dean, Koray Kavukcuoglu, Demis Hassabis, Zoubin Ghahramani, Douglas Eck, Joelle Barral, Fernando Pereira, Eli Collins, Armand Joulin, Noah Fiedel, Evan Senter, Alek Andreev, and Kathleen Kenealy. 2024.
\newblock \href {http://arxiv.org/abs/2403.08295} {Gemma: Open models based on gemini research and technology}.

\bibitem[{Tunstall et~al.(2023)Tunstall, Beeching, Lambert, Rajani, Rasul, Belkada, Huang, von Werra, Fourrier, Habib, Sarrazin, Sanseviero, Rush, and Wolf}]{z}
Lewis Tunstall, Edward Beeching, Nathan Lambert, Nazneen Rajani, Kashif Rasul, Younes Belkada, Shengyi Huang, Leandro von Werra, Clémentine Fourrier, Nathan Habib, Nathan Sarrazin, Omar Sanseviero, Alexander~M. Rush, and Thomas Wolf. 2023.
\newblock \href {http://arxiv.org/abs/2310.16944} {Zephyr: Direct distillation of lm alignment}.

\bibitem[{Wang et~al.(2023{\natexlab{a}})Wang, Liang, Meng, Sun, Shi, Li, Xu, Qu, and Zhou}]{wanggpt}
Jiaan Wang, Yunlong Liang, Fandong Meng, Zengkui Sun, Haoxiang Shi, Zhixu Li, Jinan Xu, Jianfeng Qu, and Jie Zhou. 2023{\natexlab{a}}.
\newblock \href {https://doi.org/10.18653/v1/2023.newsum-1.1} {Is {C}hat{GPT} a good {NLG} evaluator? a preliminary study}.
\newblock In \emph{Proceedings of the 4th New Frontiers in Summarization Workshop}, pages 1--11, Singapore. Association for Computational Linguistics.

\bibitem[{Wang et~al.(2018)Wang, Wang, Jia, and Yin}]{wang2018explainable}
Nan Wang, Hongning Wang, Yiling Jia, and Yue Yin. 2018.
\newblock \href {https://doi.org/10.1145/3209978.3210010} {Explainable recommendation via multi-task learning in opinionated text data}.
\newblock In \emph{The 41st International ACM SIGIR Conference on Research \& Development in Information Retrieval}, SIGIR '18, page 165–174, New York, NY, USA. Association for Computing Machinery.

\bibitem[{Wang et~al.(2023{\natexlab{b}})Wang, Zhang, Sun, and Meng}]{gcre2023}
Yequan Wang, Hengran Zhang, Aixin Sun, and Xuying Meng. 2023{\natexlab{b}}.
\newblock \href {https://api.semanticscholar.org/CorpusID:257532399} {Gcre-gpt: A generative model for comparative relation extraction}.
\newblock \emph{ArXiv}, abs/2303.08601.

\bibitem[{Wei et~al.(2023)Wei, Wang, Schuurmans, Bosma, Ichter, Xia, Chi, Le, and Zhou}]{wei2023}
Jason Wei, Xuezhi Wang, Dale Schuurmans, Maarten Bosma, Brian Ichter, Fei Xia, Ed~Chi, Quoc Le, and Denny Zhou. 2023.
\newblock \href {http://arxiv.org/abs/2201.11903} {Chain-of-thought prompting elicits reasoning in large language models}.

\bibitem[{Wolf et~al.(2020)Wolf, Debut, Sanh, Chaumond, Delangue, Moi, Cistac, Rault, Louf, Funtowicz, Davison, Shleifer, von Platen, Ma, Jernite, Plu, Xu, Le~Scao, Gugger, Drame, Lhoest, and Rush}]{hf}
Thomas Wolf, Lysandre Debut, Victor Sanh, Julien Chaumond, Clement Delangue, Anthony Moi, Pierric Cistac, Tim Rault, Remi Louf, Morgan Funtowicz, Joe Davison, Sam Shleifer, Patrick von Platen, Clara Ma, Yacine Jernite, Julien Plu, Canwen Xu, Teven Le~Scao, Sylvain Gugger, Mariama Drame, Quentin Lhoest, and Alexander Rush. 2020.
\newblock \href {https://doi.org/10.18653/v1/2020.emnlp-demos.6} {Transformers: State-of-the-art natural language processing}.
\newblock In \emph{Proceedings of the 2020 Conference on Empirical Methods in Natural Language Processing: System Demonstrations}, pages 38--45, Online. Association for Computational Linguistics.

\bibitem[{Yang et~al.(2024)Yang, Yang, Hui, Zheng, Yu, Zhou, Li, Li, Liu, Huang, Dong, Wei, Lin, Tang, Wang, Yang, Tu, Zhang, Ma, Yang, Xu, Zhou, Bai, He, Lin, Dang, Lu, Chen, Yang, Li, Xue, Ni, Zhang, Wang, Peng, Men, Gao, Lin, Wang, Bai, Tan, Zhu, Li, Liu, Ge, Deng, Zhou, Ren, Zhang, Wei, Ren, Liu, Fan, Yao, Zhang, Wan, Chu, Liu, Cui, Zhang, Guo, and Fan}]{q}
An~Yang, Baosong Yang, Binyuan Hui, Bo~Zheng, Bowen Yu, Chang Zhou, Chengpeng Li, Chengyuan Li, Dayiheng Liu, Fei Huang, Guanting Dong, Haoran Wei, Huan Lin, Jialong Tang, Jialin Wang, Jian Yang, Jianhong Tu, Jianwei Zhang, Jianxin Ma, Jianxin Yang, Jin Xu, Jingren Zhou, Jinze Bai, Jinzheng He, Junyang Lin, Kai Dang, Keming Lu, Keqin Chen, Kexin Yang, Mei Li, Mingfeng Xue, Na~Ni, Pei Zhang, Peng Wang, Ru~Peng, Rui Men, Ruize Gao, Runji Lin, Shijie Wang, Shuai Bai, Sinan Tan, Tianhang Zhu, Tianhao Li, Tianyu Liu, Wenbin Ge, Xiaodong Deng, Xiaohuan Zhou, Xingzhang Ren, Xinyu Zhang, Xipin Wei, Xuancheng Ren, Xuejing Liu, Yang Fan, Yang Yao, Yichang Zhang, Yu~Wan, Yunfei Chu, Yuqiong Liu, Zeyu Cui, Zhenru Zhang, Zhifang Guo, and Zhihao Fan. 2024.
\newblock \href {http://arxiv.org/abs/2407.10671} {Qwen2 technical report}.

\bibitem[{Yang et~al.(2022)Yang, Wang, Cai, Deng, and Wang}]{comparative2022}
Aobo Yang, Nan Wang, Renqin Cai, Hongbo Deng, and Hongning Wang. 2022.
\newblock \href {https://doi.org/10.1145/3485447.3512031} {Comparative explanations of recommendations}.
\newblock In \emph{Proceedings of the ACM Web Conference 2022}. ACM.

\bibitem[{Yang et~al.(2021)Yang, Wang, Deng, and Wang}]{yang2021explanation}
Aobo Yang, Nan Wang, Hongbo Deng, and Hongning Wang. 2021.
\newblock \href {https://doi.org/10.1145/3437963.3441726} {Explanation as a defense of recommendation}.
\newblock In \emph{Proceedings of the 14th ACM International Conference on Web Search and Data Mining}, WSDM '21, page 1029–1037, New York, NY, USA. Association for Computing Machinery.

\end{thebibliography}

\appendix
\section{QF-CES Dimensions}\label{cex_metrics}
We define \textsc{QF-CES} evaluation dimensions as follows:
\begin{enumerate}
    \item {\tt \textbf{clarity}} \textbf{(CL)}- Clarity measures the degree to which the information in the Comparative Summary is clearly presented, avoiding ambiguity and ensuring that comparisons are easy to understand. The summary should be clear, concise, and easy to comprehend, using simple language and avoiding technical jargon whenever possible. It should be well-structured and well-organized, presenting comparison of the three products in a straightforward manner. The metric evaluates the readability of the entire summary, ensuring it is free from grammatical errors and has a logical flow between different sections and points. Additionally, the clarity of the tabular data is assessed to ensure it clearly conveys the comparisons between three products.
    \item {\tt \textbf{faithfulness}} \textbf{(FL)}- Faithfulness measures the degree to which the information presented in the Comparative Summary is accurate, verifiable, and directly supported by the input data. The Comparative Summary must faithfully represent the content provided, ensuring that all details, including the query and attributes of each product are correct and inferred directly from the input. Comparative Summary will be penalized for any information that cannot be verified from the input data or if they make broad generalizations that are not supported by the input data.
    \item {\tt \textbf{informativeness}} \textbf{(IF)}- Informativeness evaluates the extent to which the Comparative Summary comprehensively covers all relevant aspects and attributes of the products being compared. This metric assesses the presence and completeness of essential attributes and features in the comparison, including the product title, base price, final price, key attributes dynamically selected from the product opinion summaries, pros, cons, and average rating. The summary should ensure that all majorly discussed aspects are covered and any missing values are properly marked as "N/A". Summaries should be penalized for missing significant aspects and rewarded for thorough coverage of the aspects from the provided information.
    \item {\tt \textbf{format adherence}} \textbf{(FoA)}-  This metric evaluates the extent to which the Comparative Summary follows the prescribed format. The Comparative Summary should consist of two main parts: (\textit{1}) A tabular comparison of the three products. (\textit{2}) A final verdict summary.
    
    The tabular comparison should list products in columns and attributes in rows,including dynamically selected attributes based on the user query and essential attributes such as Base Price, Final Price, Average Rating, Pros, and Cons. It verifies that dynamically selected attributes are appropriately named and not using placeholders. The final verdict summary should provide a concise overview of the comparison among three products. The metric assesses the presence, completeness, and proper formatting of both these components (the tabular comparison along with the final verdict), as well as the overall organization and consistency of the entire summary.
    
    \item {\tt \textbf{query relevance}} \textbf{(QR)}- This metric evaluates how well the Comparative Summary addresses the user's query. It assesses two main components: (\textit{1}) \textit{The tabular comparison:} Ensures that only the most relevant information and dynamic attributes are present, directly addressing the user query without including irrelevant details. (\textit{2}) \textit{The final verdict summary:} Verifies that the user query is explicitly addressed, providing clear suggestions that enable the user to make an informed buying decision.
    
    The metric measures the overall relevance and usefulness of the Comparative Summary in helping the user make an informed decision based on their specific query.
\end{enumerate}

\section{M-OS Evaluation}\label{mos}
 Figure \ref{fig:opsum} represents the model-wise performance across $7$ dimensions: {\fl} (FL), {\coh} (CO), {\asp} (AC), {\fa} (FF), {\rel} (RL), {\sent} (SC), and {\spec} (SP). The scores are given by {\llamainstruct{3.1}{70}} as evaluator, for \verb|n=100| generations.
\begin{figure}[htp]
    \centering
    \includegraphics[width=1\columnwidth]{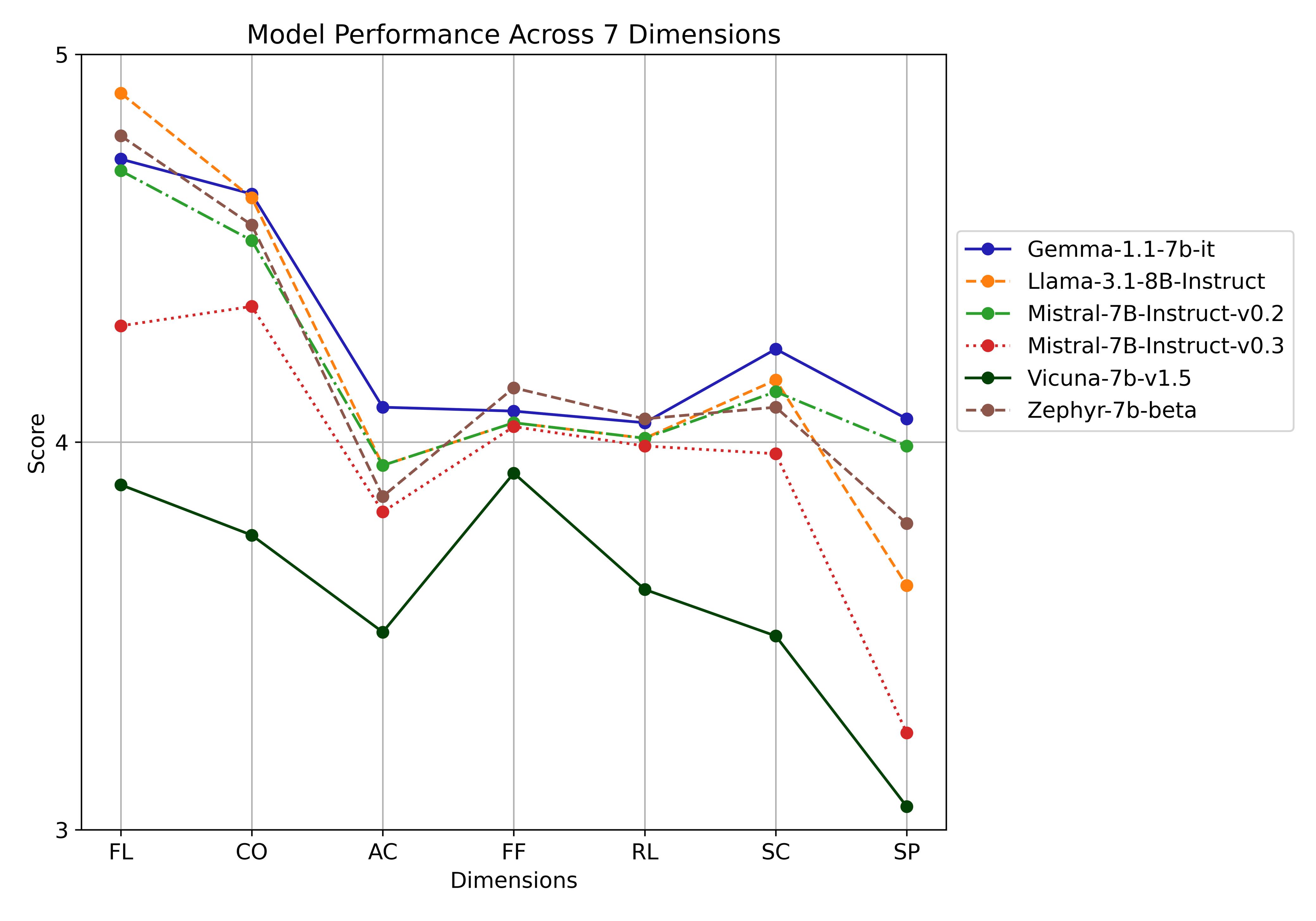}
    \caption{Various open-source models performance across $7$ dimensions by {\llamainstruct{3.1}{70}} as evaluator } 
    \label{fig:opsum}
\end{figure}

\section{QF-CES Evaluation}
Figure \ref{fig:different_evaluators} represents model-wise averaged score given by various LLM as evaluators of \textsc{QF-CES} along $5$ dimensions: {\cl} (CL), {\fa} (FA), {\ifo} (IF), {\foa} (FoA) and {\qr} (QR)
\begin{figure}[htp]
    \centering
    \includegraphics[width=1\columnwidth]{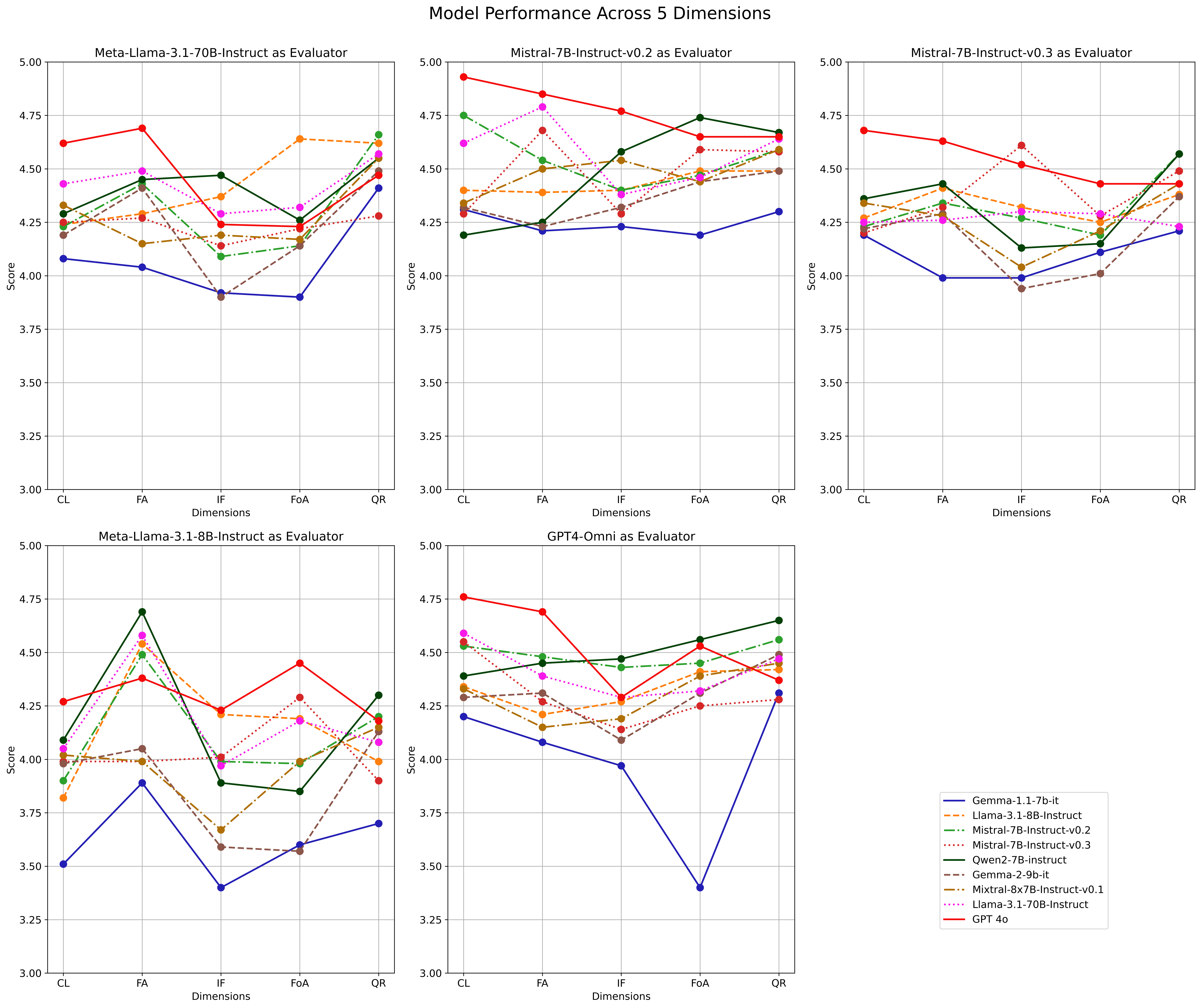}
    \caption{Performance of different models as rated by human annotators (Round-II). We observe that {\gptfourstandard} performs the best followed by {\qweninstruct{2}}.} 
    \label{fig:different_evaluators}
\end{figure}

\section{LLM as Evaluators}
Figure \ref{fig:cex_human} represents model-wise averaged annotator ratings of \textsc{QF-CES} along $5$ dimensions: {\cl} (CL), {\fa} (FA), {\ifo} (IF), {\foa} (FoA) and {\qr} (QR)
\begin{figure}[htp]
    \centering
    \includegraphics[width=1\columnwidth]{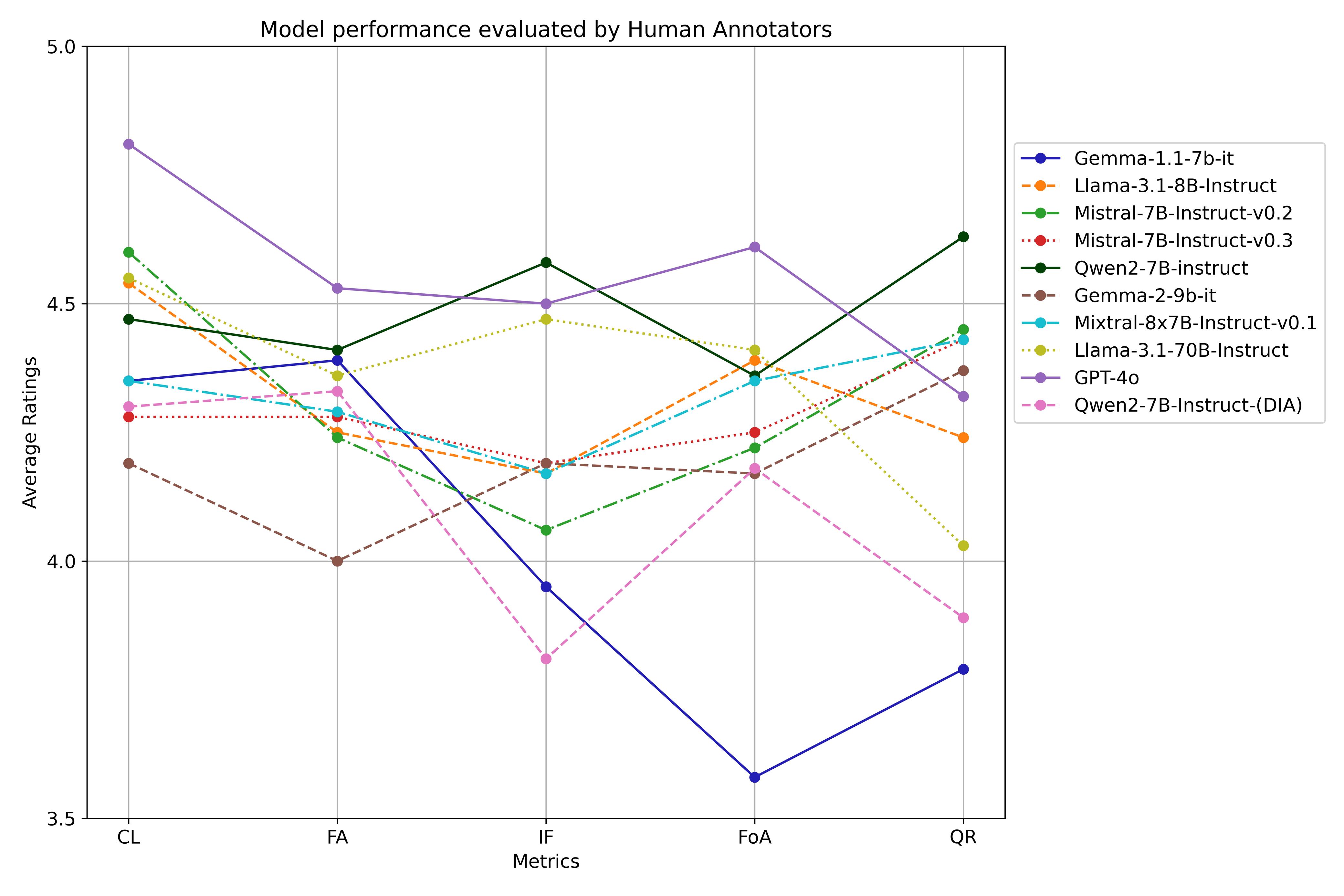}
    \caption{Performance of different models as rated by human annotators (Round-II). We observe that {\gptfourstandard} performs the best followed by {\qweninstruct{2}}.} 
    \label{fig:cex_human}
\end{figure}



\section{Rater Agreement}\label{rater_agreement} Table \ref{tab:rater_correlation} reports the pairwise correlations between raters as well as the correlation between each rater and average ratings for both Round-I and Round-II.
\begin{table*}[htp]
    \centering
    \small
    \begin{tabular}{clcccccccccc}
    \toprule
          & & \multicolumn{2}{c}{\textbf{CL} $\uparrow$}  & \multicolumn{2}{c}{\textbf{FA} $\uparrow$} & \multicolumn{2}{c}{\textbf{IF} $\uparrow$} & \multicolumn{2}{c}{\textbf{FoA} $\uparrow$} & \multicolumn{2}{c}{\textbf{QR} $\uparrow$} \\
         \cmidrule(lr){3-4} \cmidrule(lr){5-6} \cmidrule(lr){7-8} \cmidrule(lr){9-10} \cmidrule(lr){11-12}
         & & $\rho$ & $\tau$ & $\rho$ & $\tau$ & $\rho$ & $\tau$ & $\rho$ & $\tau$ & $\rho$ & $\tau$ \\  
   
    \midrule
        \multicolumn{12}{c}{\textit{Pairwise correlation among raters}}\\
    \cmidrule{2-12}
        \multirow{6}{*}{\rotatebox[origin=c]{90}{\textbf{Round-I}}} 
        & \textbf{A1-A2} & $0.59$ & $0.59$ & $0.59$ & $0.57$ & $0.49$ & $0.47$ & $0.53$ & $0.51$ & $0.39$ & $0.37$ \\
        & \textbf{A2-A3} & $0.58$ & $0.57$ & $0.58$ & $0.57$ & $0.33$ & $0.31$ & $0.46$ & $0.45$ & $0.32$ & $0.29$ \\
        & \textbf{A1-A3} & $0.60$ & $0.60$ & $0.57$ & $0.56$ & $0.38$ & $0.36$ & $0.58$ & $0.56$ & $0.47$ & $0.44$ \\
    \cmidrule{2-12}
        & \textbf{AVG-I} & $0.59$ & $0.59$ & $0.58$ & $0.57$ & $0.40$ & $0.38$ & $0.52$ & $0.51$ & $0.39$ & $0.37$ \\
    \cmidrule{2-12}
        \multicolumn{12}{c}{\textit{Pairwise correlation between raters and the overall average ratings}}\\
    \cmidrule{2-12}
        & \textbf{A-A1} & $0.82$ & $0.78$ & $0.86$ & $0.80$ & $0.74$ & $0.68$ & $0.83$ & $0.77$ & $0.79$ & $0.72$ \\
        & \textbf{A-A2} & $0.84$ & $0.79$ & $0.79$ & $0.74$ & $0.78$ & $0.72$ & $0.79$ & $0.73$ & $0.72$ & $0.64$ \\
        & \textbf{A-A3} & $0.84$ & $0.80$ & $0.83$ & $0.78$ & $0.73$ & $0.67$ & $0.81$ & $0.75$ & $0.78$ & $0.71$ \\
    \cmidrule{2-12}
        & \textbf{AVG-II} & $0.83$ & $0.79$ & $0.83$ & $0.77$ & $0.75$ & $0.69$ & $0.81$ & $0.75$ & $0.76$ & $0.69$ \\
        
    \midrule
        \multicolumn{12}{c}{\textit{Pairwise correlation among raters}}\\
    \cmidrule{2-12}
        \multirow{6}{*}{\rotatebox[origin=c]{90}{\textbf{Round-II}}} 
        & \textbf{A1-A2} & $0.80$ & $0.80$ & $0.80$ & $0.79$ & $0.77$ & $0.76$ & $0.81$ & $0.80$ & $0.85$ & $0.83$ \\
        & \textbf{A2-A3} & $0.78$ & $0.78$ & $0.79$ & $0.79$ & $0.77$ & $0.75$ & $0.78$ & $0.77$ & $0.72$ & $0.70$ \\
        & \textbf{A1-A3} & $0.78$ & $0.78$ & $0.82$ & $0.81$ & $0.75$ & $0.73$ & $0.81$ & $0.80$ & $0.70$ & $0.68$ \\
    \cmidrule{2-12}
        & \textbf{AVG-I} & $0.79$ & $0.78$ & $0.80$ & $0.80$ & $0.76$ & $0.75$ & $0.80$ & $0.79$ & $0.75$ & $0.74$ \\
    \cmidrule{2-12}
    \multicolumn{12}{c}{\textit{Pairwise correlation between raters and the overall average ratings}}\\
    \cmidrule{2-12}
        & \textbf{A-A1} & $0.91$ & $0.87$ & $0.92$ & $0.88$ & $0.86$ & $0.82$ & $0.92$ & $0.88$ & $0.90$ & $0.85$ \\
        & \textbf{A-A2} & $0.92$ & $0.87$ & $0.91$ & $0.87$ & $0.88$ & $0.84$ & $0.92$ & $0.88$ & $0.92$ & $0.87$ \\
        & \textbf{A-A3} & $0.92$ & $0.87$ & $0.94$ & $0.89$ & $0.91$ & $0.87$ & $0.92$ & $0.88$ & $0.88$ & $0.83$ \\
    \cmidrule{2-12}
        & \textbf{AVG-II} & $0.92$ & $0.87$ & $0.92$ & $0.88$ & $0.89$ & $0.85$ & $0.92$ & $0.88$ & $0.90$ & $0.85$ \\
    \bottomrule
    \end{tabular}
    \caption{\textbf{Rater Correlations:} Pairwise \textit{Spearman} ($\rho$) and \textit{Kendall Tau} ($\tau$) correlations at summary-level for 3 raters A1, A2, and A3 along with the average of their ratings.}
    \label{tab:rater_correlation}
\end{table*}

\end{document}